\documentclass[10pt,twocolumn,letterpaper]{article}

\usepackage{iccv}
\usepackage{times}
\usepackage{epsfig}
\usepackage{graphicx}
\usepackage{amsmath}
\usepackage{amssymb}

\usepackage{authblk}
\usepackage{times}
\usepackage{epsfig}
\usepackage{graphicx}
\usepackage{enumerate}
\usepackage[linesnumbered,ruled]{algorithm2e}
\usepackage{bm}
\usepackage{color}
\usepackage{bbm}
\usepackage{amsthm}
\usepackage{url}
\usepackage{subfigure}
\usepackage{multirow}
\usepackage{booktabs}
\usepackage{balance}

\newcommand{\peter}[1]{#1}

\usepackage[breaklinks=true,bookmarks=false]{hyperref}

\newtheorem{definition}{Definition}
\newtheorem{theorem}{Theorem}
\newtheorem{example}{Example}

\newcommand{\argmax}{\operatornamewithlimits{argmax}}
\newcommand{\argmin}{\operatornamewithlimits{argmin}}

\iccvfinalcopy % *** Uncomment this line for the final submission

\def\iccvPaperID{****} % *** Enter the ICCV Paper ID here

\ificcvfinal\fi

\begin{document}

%%%%%%%%% TITLE
\title{Finding Representative Interpretations on Convolutional Neural Networks}

% \author{Peter Cho-Ho Lam\\
% Huawei Canada Technologies Co. Ltd.\\
% {\tt\small cho.ho.lam@huawei.com}
% \and
% Lingyang Chu\\

\author[1]{Peter Cho-Ho Lam}
\author[2]{Lingyang Chu\thanks{Peter Cho-Ho Lam and Lingyang Chu contribute equally in this work. 
Please refer to \cite{lam2021finding} for the complete version of this paper.
%The code of this work is accessible at 
%\protect\url{https://lingyangchu.github.io}
%\protect\url{https://t.ly/k8k1}
}
}
\author[1]{Maxim Torgonskiy}
\author[3]{Jian Pei}
\author[1]{Yong Zhang}
\author[1]{Lanjun Wang}
\affil[1]{Huawei Canada Technologies Co., Ltd.}
\affil[2]{McMaster University}
\affil[3]{Simon Fraser University}
\affil[ ]{\textit {\{cho.ho.lam, maxim.torgonskiy, yong.zhang3, lanjun.wang\}@huawei.com}}
\affil[ ]{\textit {chul9@mcmaster.ca}}
\affil[ ]{\textit {jpei@cs.sfu.ca}}

\maketitle
\ificcvfinal\thispagestyle{empty}\fi

%%%%%%%%% ABSTRACT
\begin{abstract}
Interpreting the decision logic behind effective deep convolutional neural networks (CNN) on images complements the success of deep learning models. However, the existing methods can only interpret some specific decision logic on individual or a small number of images.  To facilitate human understandability and generalization ability, it is important to develop representative interpretations that interpret common decision logics of a CNN on a large group of similar images, which reveal the common semantics data contributes to many closely related predictions.  In this paper, we develop a novel unsupervised approach to produce a highly representative interpretation for a large number of similar images. We formulate the problem of finding representative interpretations as a co-clustering problem, and convert it into a submodular cost submodular cover problem based on a sample of the linear decision boundaries of a CNN.  We also present a visualization and similarity ranking method. Our extensive experiments demonstrate the excellent performance of our method.
\end{abstract}

%%%%%%%%% BODY TEXT
\section{Introduction}
\label{sec:intro}

Interpretability becomes more and more important for machine learning~\cite{molnar2019, ribeiro2016should}. As a fundamental deep learning model, the interpretability of convolutional neural networks (CNNs)~\cite{he2016deep, krizhevsky2012imagenet} has been intensively explored~\cite{chattopadhay2018grad, omeiza2019smooth, ramaswamy2020ablation, selvaraju2017grad, wang2020score, zhou2016learning}.  For example, many interpretation methods~\cite{chu2018exact, ribeiro2016should, selvaraju2017grad, Srinivas2019FullGradientRF} have been proposed to interpret the specific decision logic of a CNN on a single input image or a small group of images.  However, the challenge of interpretation on CNNs for image classification is still far from being settled.  An interpretation accommodating only one individual image or a small group of images is highly sensitive to the noise contained in the interpreted image~\cite{ghorbani2019interpretation}, and thus cannot be robustly reused to interpret the predictions on a large group of unseen images~\cite{alvarez2018robustness, Dombrowski2019ExplanationsCB, subramanya2019fooling}.

To tackle the challenge, we advocate seeking \emph{representative interpretations}~\cite{carvalho2019machine}, which interpret the common decision logic of a CNN.  
%Representative interpretations can capture the common semantics of a large group of similar images, and generalize well in making predictions on unseen images~\cite{kim2018interpretability, zhang2019interpreting}. 
A representative interpretation is more convincing, because it can be validly reused to interpret the same decision logic that governs the predictions on a large number of unseen images~\cite{carvalho2019machine}. Moreover, a representative interpretation also provides a deeper insight into the common semantic of a large number of similar images~\cite{Ghorbani2019TowardsAC, kim2018interpretability, zhang2019interpreting}. Such common semantics are generally believed to be captured by CNNs to achieve excellent prediction performance~\cite{fong2018net2vec, zhou2016learning,zhou2018interpretable}.

However, interpreting the common decision logic of a CNN on a large group of similar images in an unsupervised manner is a novel problem that has not been systematically studied in literature~\cite{Ghorbani2019TowardsAC}.  Can we straight-forwardly extend the existing interpretation methods to produce representative interpretations on a CNN? Unfortunately, many existing methods~\cite{chattopadhay2018grad,chu2018exact, omeiza2019smooth, selvaraju2017grad,ribeiro2016should} focus on interpreting the specific decision logic of a CNN on a single input image or a small group of images, and thus cannot produce representative interpretations. More critically, it is difficult to produce a representative interpretation without knowing a large group of similar images that are predicted by a common decision logic of a CNN. As discussed in Section~\ref{sec:rw}, some concept-based interpretation methods summarize the common decision-making concepts from a group of conceptually similar images. However, those methods are either limited to sophisticatedly customized CNN models~\cite{chen2018looks, zhang2019interpreting}, or require a large set of conceptually annotated images~\cite{fong2018net2vec, zhou2018interpretable}
%\peter{(Removed TCAV since a reviewer disagrees)} 
that are too expensive to obtain in most real world applications~\cite{Ghorbani2019TowardsAC}.

In this paper, we propose the novel unsupervised task of finding representative interpretations on convolutional neural networks. The goal is to find and visualize the common decision logic of a CNN that governs the predictions on an input image and a large set of similar images.  We make a series of technical contributions as follows.

We model the common decision logic of a CNN on a group of similar images as a decision region formed by a set of pieces of the decision boundaries of the CNN. All images contained in this decision region are predicted as the same class by the same decision logic characterized by the decision boundary pieces. We formulate our task as a co-clustering problem to simultaneously find the largest group of similar images and the corresponding decision region containing them. We further convert the co-clustering problem into an NP-hard submodular cost submodular cover problem, and develop an efficient heuristic method to solve it without requiring any conceptual image annotations or any modifications to CNNs.
To produce understandable interpretations for the images contained in a decision region, we visualize the boundaries of the decision region as heat maps to identify important image regions for predictions. We also rank the similar images according to their semantic distances to the input image.  Last, we conduct extensive experiments to examine the interpretation quality and reusability of our representative interpretations.

%------------------------------------------------------------------------

\section{Related Works}
\label{sec:rw}

Our work is related to the four major types of existing interpretation methods for CNNs in image classification.

The \textbf{gradient-based methods}~\cite{chattopadhay2018grad, omeiza2019smooth, selvaraju2017grad, simonyan2013deep, zeiler2014visualizing, zeiler2011adaptive} produce interpretations on an input image by computing and visualizing different types of gradients.
The early studies~\cite{adebayo2018local, levine2019certifiably, lu2020robust, simonyan2013deep, smilkov2017smoothgrad, springenberg2014striving, Srinivas2019FullGradientRF, sundararajan2017axiomatic, zeiler2014visualizing, zeiler2011adaptive} straight-forwardly visualize the gradient of the score of a predicted class with respect to an input image. 
%However, interpretations produced by these methods usually consist of noisy scattered pixels that are difficult to understand~\cite{sundararajan2019exploring}.
The class activation mapping (CAM) methods~\cite{chattopadhay2018grad, omeiza2019smooth, ramaswamy2020ablation, selvaraju2017grad, wang2020score, zhou2016learning}
use discriminative localization to produce continuous image regions as interpretations.
The relevance score methods~\cite{bach2015pixel, montavon2017explaining, nam2020relative, shrikumar2017learning} produce interpretations by back-propagating a set of relevance scores with respect to an input image.

All those methods focus on producing interpretations that are highly correlated to the prediction on a single input image.
However, such interpretations are not representative, because they are highly sensitive to the noise contained in the images that are interpreted~\cite{ghorbani2019interpretation}. Thus, such interpretations cannot be effectively reused to interpret the predictions on a large group of unseen images~\cite{alvarez2018robustness, Dombrowski2019ExplanationsCB, subramanya2019fooling}.

The \textbf{model approximation
methods}~\cite{chu2018exact, frosst2017distilling, ibrahim2019global, lundberg2017unified, ribeiro2016should, ribeiro2018anchors, tan2018learning, yang2018global} produce interpretations on a deep neural network by approximating it locally or globally with an interpretable agent model, such as decision trees~\cite{frosst2017distilling} and linear classifiers~\cite{chu2018exact, ribeiro2016should}.

The local approximation methods, such as Openbox~\cite{chu2018exact}, LIME~\cite{ribeiro2016should} and Anchors~\cite{ribeiro2018anchors}, produce interpretations by approximating the predictions of a neural network in a local similarity neighbourhood of a target image for interpretation.
The interpretations produced by Openbox~\cite{chu2018exact} and LIME~\cite{ribeiro2016should} are only applicable to a very small neighbourhood, which largely limits their ability to produce representative interpretations~\cite{ribeiro2018anchors}.
Interpretations produced by Anchors~\cite{ribeiro2018anchors} are applicable to a large neighbourhood. However, Anchors cannot produce effective interpretations for CNNs on images, because the applicable neighbourhoods for Anchors are not well-defined on images~\cite{molnar2019}.

The global approximation methods~\cite{frosst2017distilling, ibrahim2019global, tan2018learning, yang2018global} attempt to approximate the entire neural network with an interpretable agent model. 
These methods intend to produce interpretations that apply to multiple similar images. 
However, the agent models are usually too simple to
accurately approximate complex deep neural networks~\cite{chu2018exact}.
Therefore, most of these methods perform poorly on modern CNNs trained on complicated natural images \cite{wan2020nbdt}.

The \textbf{conceptual interpretation methods}~\cite{fong2018net2vec, Ghorbani2019TowardsAC, kim2018interpretability, zhang2018interpreting, zhang2017growing, zhang2019interpreting, zhou2018interpretable} aim to identify a set of concepts that contribute to the predictions on a pre-defined group of conceptually similar images. Here, a concept is either supervision information or learned from data using another model.
TCAV~\cite{kim2018interpretability}, Net2Vec~\cite{fong2018net2vec} and IBD~\cite{zhou2018interpretable} interpret a neural network's internal state in terms of human-friendly concepts.
These methods require a dataset %\peter{(data set becomes dataset)} 
with high-quality conceptual labels, which is difficult, if possible at all, to obtain in real world applications~\cite{Ghorbani2019TowardsAC}.
Some methods~\cite{zhang2019interpreting} use decision trees to learn concepts without using a dataset 
%\peter{(data set becomes dataset)} 
with conceptual labels.
The learned concepts reveal the common decision logic of a group of similar images.
However, these methods require sophisticated customization on deep neural networks, and thus suffer from major limitations in applicability to general CNNs.

ACE~\cite{Ghorbani2019TowardsAC} uses a set of images in the same class to summarize important concepts by clustering similar image segments and removing outlying image segments.
A subset of the summarized concepts can be straight-forwardly used to interpret the predictions on an input image. 
However, as shown in Section~\ref{sec:cs}, ACE cannot produce representative interpretations because its interpretations on similar images are quite different due to the high sensitivity of the image segmentation method and the clustering process.

The \textbf{example-based methods}~\cite{charpiat2019input, goyal2019counterfactual, khanna2019interpreting, koh2017understanding, wu2020smint, wu2018sharing} find exemplar images to interpret the decision of a deep neural network on a single input image. 
The exemplar images are similar to the input image, but they do not provide feature level interpretations to reveal the decision logic on the input image or the exemplar images~\cite{molnar2019}.
Therefore, there is no guarantee that the exemplar images are predicted by the same decision logic as the input image.

\peter{Instead of finding explanatory example for each single input, the \textbf{prototype-based methods}~\cite{bien2011prototype, kim2016examples} summarize the entire model using a small number of instances as prototypes. %These instances can also serve as a classifier by using the nearest neighbor algorithm. 
%This type of outputs is very close to achieving a representative interpretation;
However, the selection of prototypes considers very little about the decision process of the model. Thus they do not provide much insight on the model's decision logic.}

As discussed in Section~\ref{sec:intro}, the existing methods cannot produce representative interpretations for CNNs on image classification, which is the challenge tackled by this paper. 

%------------------------------------------------------------------------

\section{Problem Formulation}
\label{sec:pd}

In this section, we first discuss the idea of representative interpretations for image classification by CNN~\cite{he2016deep, krizhevsky2012imagenet}. 
Then, we formulate the problem of finding representative interpretation on CNNs.
%Last, we provide an intuitive example to illustrate the key idea of our method.

\subsection{Representative Interpretations}

In this paper, we consider image classification using CNNs that adopt piecewise linear activation functions, such as MaxOut~\cite{goodfellow2013maxout} and the family of ReLU~\cite{glorot2011deep, he2015delving, nair2010rectified}, for hidden neurons. 
Denote by $\mathcal{X}$ the space of input images.  Let $F:\mathcal{X}\rightarrow \mathbb{R}^C$ be a trained CNN to be interpreted and $C$ the number of classes of images.  We assume a set of unlabeled reference images $R\subseteq \mathcal{X}$  that are used to generate representative interpretations.

For in input image $x\in\mathcal{X}$, $F$ produces a vector of prediction scores
$%\begin{equation}\nonumber
F(x)=\left[F_1(x), F_2(x), \ldots, F_C(x)\right],
$ %\end{equation}
where $F_i(x)$ is the predicted score of the $i$-th class before the normalization of the last softmax layer.
The predicted class of $x$ is $Class(x) = \argmax_i F_i(x)$.

Roughly speaking, an interpretation of a classification model is a description of the logics used by the model to produce the classification.  While different approaches may employ different representations for interpretations, in image classification, for example, a common way to represent an interpretation for an image is a heat map that identifies the important regions leading to the prediction~\cite{chattopadhay2018grad, omeiza2019smooth, ramaswamy2020ablation, selvaraju2017grad, wang2020score, zhou2016learning}.  Since interpretation is often related to generalization, an important idea to make interpretation more understandable is to seek for representative interpretations that are applicable to not only a specific image but also many similar images~\cite{carvalho2019machine}.  

\begin{definition}
For an input image $x\in\mathcal{X}$ predicted as $Class(x)$ by $F$, a \textbf{representative interpretation} on $x$ is an interpretation that reveals the common decision logic of $F$ on $x$ as well as a large set of reference images in $R$.
\end{definition}

In order to construct representative interpretations, let us first consider how to model the decision logic of $F$.

For an input image $x\in \mathcal{X}$, denote by $\psi(x)$ the feature map produced by the last convolutional layer of $F$ and by $\Omega=\{\psi(x) \mid x\in\mathcal{X}\}$ the space of feature maps. Following the existing studies~\cite{chattopadhay2018grad, ramaswamy2020ablation, selvaraju2017grad, wang2020score, zhang2019interpreting}, we model the decision logic of $F$ on $x$ by analyzing how the fully connected layers of $F$, denoted by $G:\Omega\rightarrow \mathbb{R}^C$, use the feature map $\psi(x)$ to make the prediction on $x$.
For any input image $x$, we have $G(\psi(x)) = F(x)$.

Since $F$ adopts a piecewise linear activation function in all the hidden neurons, 
the decision logic of $G$ can be characterized by a piecewise linear decision boundary that consists of connected pieces of decision hyperplanes~\cite{montufar2014number, raghu2017expressive}. 
We call these hyperplanes \textbf{linear boundaries}, and denote by $\mathcal{P}$ the set of the linear boundaries of $G$.

The linear boundaries in $\mathcal{P}$ partition the space of feature maps $\Omega$ into a large number of convex polytopes~\cite{montufar2014number, raghu2017expressive}, where each convex polytope defines a decision region that predicts all the images contained in the region to be the same class.
The linear decision boundaries of the convex polytopes encode the decision logic of the fully connected layers of $F$~\cite{chu2018exact}.

Inspired by the above insights, we model the decision logic of $F$ on an input image $x\in\mathcal{X}$ by a convex polytope in $\Omega$ that contains $\psi(x)$.  This convex polytope must be induced by a subset of the linear boundaries of $G$, denoted by $P(x)\subseteq \mathcal{P}$. The convex polytope defines a decision region such that the reference images contained in the region are also predicted as $Class(x)$. 

To keep our presentation concise, we overload symbol $P(x)$ to denote the decision region, that is, the convex polytope it defines. 

When the context is clear, we say a feature map $\psi(x')$ of an image $x'\in\mathcal{X}$ is \textbf{contained} in $P(x)$ if $\psi(x')$ is contained in the convex polytope defined by $P(x)$. 
We also say $x'$ is \textbf{covered} by $P(x)$ if $\psi(x')$ is contained in $P(x)$.

Since all the reference images covered by $P(x)$ are predicted as the same class as $x$ by the same set of decision boundaries, the contained images are sharing the same decision logic as $x$.  Therefore, $P(x)$ is a representative interpretation on $x$ because it reveals the common decision logic of $G$ when making predictions on $x$ as well as all the reference images covered by $P(x)$.

One issue is that the linear boundaries in $P(x)$ are in the space $\Omega$ of feature maps, and thus are not easy to comprehend by human being.  To tackle this issue,  in Section~\ref{sec:visual}, we will develop a method to visualize $P(x)$ as meaningful heat maps that identify important regions of the covered images.

According to Carvalho et al.~\cite{carvalho2019machine}, the representativeness of an interpretation is the number of images that the interpretation is applicable to. Since the decision logic modeled by $P(x)$ applies to all the images it covers, we define the \textbf{representativeness} of $P(x)$ as the number of reference images covered by $P(x)$.
%\begin{definition}
%\label{def:rep}
%The \textbf{representativeness} of $P(x)$ is the number of reference images covered by $P(x)$.
%\end{definition}

\subsection{Finding Representative Interpretations}

%Now, we define the representative interpretation finding task on convolutional neural networks (\textit{representative interpretation finding for short}), and formulate it as a co-clustering problem.

%\begin{definition}
Given a trained CNN $F$ and a set of reference images $R\subseteq \mathcal{X}$, for an input image $x\in\mathcal{X}$ whose prediction by $F$ is to be interpreted, the task of \textbf{finding representative interpretation} is to find $P(x)\subseteq \mathcal{P}$ with the largest representativeness, and visualize $P(x)$ as human comprehensible heat maps to identify image regions that are important for the prediction. 
%\end{definition}
Here, we focus on formulating a co-clustering problem that finds $P(x)$ with the largest representativeness, and leave visualization of $P(x)$ to Section~\ref{sec:visual}.

Denote by $P(x)\cap R$ the set of reference images covered by $P(x)$. $\left| P(x) \cap R \right|$, the number of reference images covered by $P(x)$, is exactly the representativeness of $P(x)$.

Denote by $D(x)=\{x'\in R \mid Class(x')\neq Class(x)\}$ the set of reference images that are predicted by $F$ to a class different from $Class(x)$.
Let $P(x) \cap D(x)$ be the set of images in $D(x)$ covered by $P(x)$.
%, and by $\left| P(x) \cap D(x) \right|$ the cardinality of this set of images. 
The task of finding representative interpretations can be formulated as the following co-clustering problem~\cite{dhillon2003information}.
\begin{subequations}\label{eq:ideal}
\begin{align}
	& \max_{P(x)\subseteq \mathcal{P}} \left| P(x) \cap R \right| \label{eq:ideal_a}\\
	& \text{s.t. } \left| P(x) \cap D(x) \right| = 0 \label{eq:ideal_b}
\end{align}
\end{subequations}
Here, the objective in Equation~\eqref{eq:ideal_a} is to maximize the representativeness of $P(x)$, and the constraint in Equation~\eqref{eq:ideal_b} requires $P(x)$ to be a decision region that covers no reference image in $R$ that is predicted to a class different from $Class(x)$.

The optimization problem in Equation~\eqref{eq:ideal} is a co-clustering problem because it simultaneously finds two clusters to optimize the representativeness of $P(x)$. One cluster is the set of similar images in $P(x)\cap R$ that are predicted by the common decision logic characterized by $P(x)$.
The other cluster is the set of linear boundaries in $P(x)$ that characterizes the decision logic of $G$ on all the images in $P(x)\cap R$. By solving the co-clustering problem, we can simultaneously find a maximal set of similar images including the input image and many reference images, as well as a representative interpretation $P(x)$ that characterizes the common decision logic on these similar images.

\begin{figure}[t]
\centering
%===================================================
\subfigure[An example of $P(x)$]{\includegraphics[height=35mm]{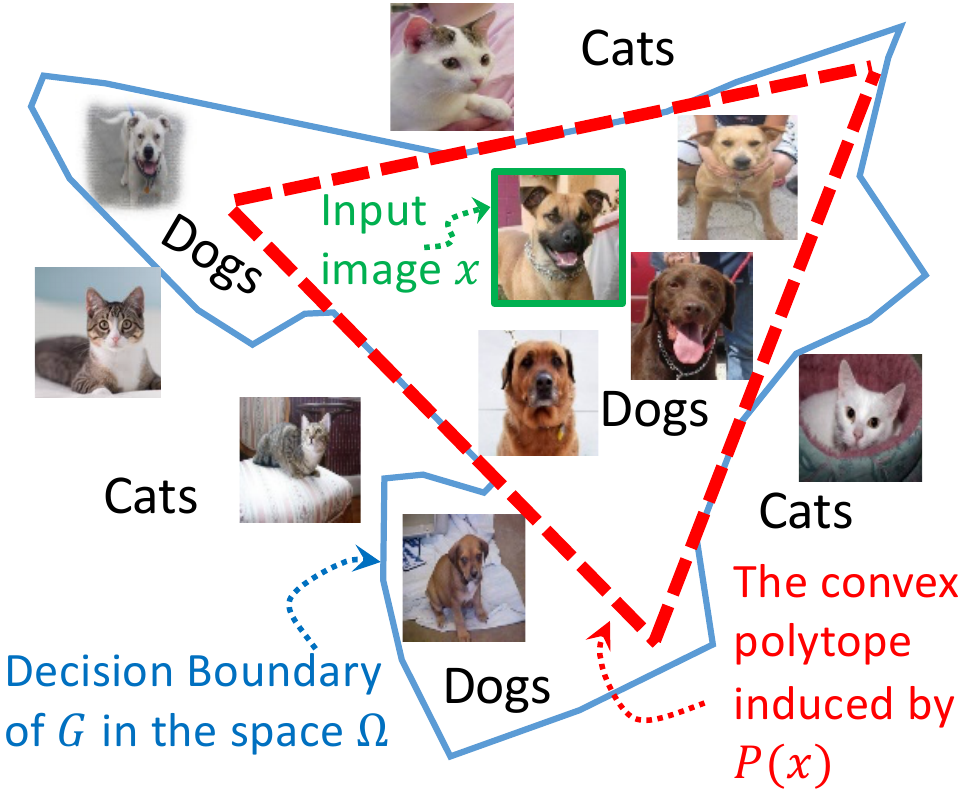}}
\hspace{2mm}
\subfigure[Visualizing $P(x)$]{\includegraphics[height=35mm]{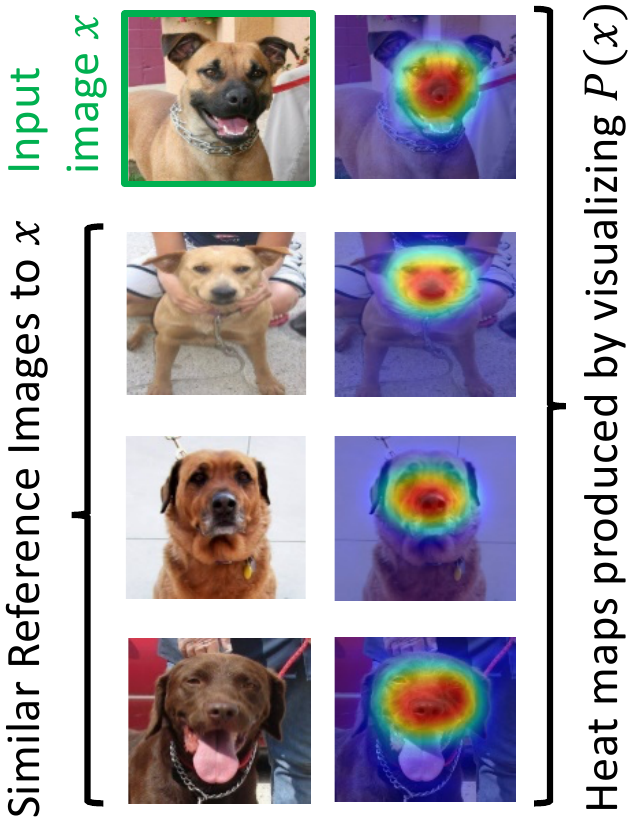}}
%===================================================
\caption{
An example showing our ideas. The input image is marked by a green box. The other images are reference images.
}
\label{fig:short}
\end{figure}

%\subsection{Intuitions and Ideas}
\begin{example}
%Let us use an example to illustrate the key idea of our method.   
In Figure~\ref{fig:short}(a), 
the blue line segments are the piecewise linear decision boundaries of $G$ in the space of $\Omega$. 
This decision boundary characterizes the decision logic of $G$ in separating the feature maps of images of dogs and those of cats. We assume $\Omega$ is a two dimensional space for simplicity.

Given the input image $x$ of a dog in the green box in the figure, 
the objective of our co-clustering problem is to find a set of linear boundaries $P(x)$, such that the convex polytope induced by $P(x)$ covers the input image $x$, and as many reference images of dogs as possible, but does not cover any reference image of cat. Following this objective, we can see that, in Figure~\ref{fig:short}(a), the optimal $P(x)$ consists of the three linear boundaries in red dashed lines, which induce a triangular convex polytope that covers the largest set of images of dogs without covering any image of cat.

The linear boundaries in $P(x)$ characterize the common decision logic of $G$ on all the images of dogs covered by $P(x)$. Since $P(x)$ covers a large number of images of dogs, the decision logic characterized by $P(x)$ is highly representative, and can be generally applied to interpret the predictions on all the images of dogs covered.
\end{example}

%\chu{Example should say about the principle}

Comparing to the existing methods that only interpret the prediction of the input image, our method shows similar reference images with conceptually similar heat maps to the input image. This makes interpretations even more effective and convincing.

%The similar reference images in Figure~\ref{fig:short}(b) are ranked based on their semantic distances to the input image $x$. We will discuss later in Section~\ref{sec:ranking} about how to use $P(x)$ to compute the semantic distance between two images.

%------------------------------------------------------------------------

\section{Finding Practical Solutions}
\label{sec:algorithm}

The optimization problem in Equation~\eqref{eq:ideal} cannot be exactly solved for large CNNs in practice, because the set of linear boundaries $\mathcal{P}$ of a complex CNN is too large to obtain, 
and we have to search the power set of $\mathcal{P}$ to find the optimal solution to the combinatorial problem in Equation~\eqref{eq:ideal}.

To tackle the challenge, we first obtain a sample $\mathcal{Q}$ of $\mathcal{P}$, and use $\mathcal{Q}$ to convert the co-clustering problem in Equation~\eqref{eq:ideal} into a submodular cost submodular cover (SCSC)  problem. 
Then, we propose a greedy algorithm to find a good solution to the SCSC problem.

\subsection{Converting into an SCSC Problem}

In this section, we first introduce how to sample a subset of linear boundaries $\mathcal{Q}$ from $\mathcal{P}$. Then, we discuss how to use $\mathcal{Q}$ to convert the co-clustering problem in Equation~\eqref{eq:ideal} into an SCSC problem.

Denote by $t \in \Omega$ a feature map, and by $\langle W, t\rangle+b=0$ a hyperplane in $\Omega$ defined by the coefficients $W$ and $b$. Here, $\langle W, t \rangle$ is the inner product between the tensors $W$ and $t$.

We can easily derive from \cite{chu2018exact} that, 
for any image $x\in\mathcal{X}$, the hyperplane induced by the following coefficients 
\begin{equation}
\label{eq:wb}
\left\{
\begin{aligned}
	& W=\frac{\partial \Delta G(t)}{\partial t}\mid_{t=\psi(x)} \\
	& b=\Delta G(t) - \langle W, t\rangle\mid_{t=\psi(x)}
\end{aligned}
\right.
\end{equation}
is exactly a linear boundary $\mathbf{h}\in\mathcal{P}$. 
\peter{Here, 
$
\Delta G(t) := {\max}_1 G(t) - {\max}_2 G(t),
$
is the difference of the largest two prediction scores; $\max_iG(t)$ is the $i$-th largest value in the vector $G(t)$.}

% \peter{Here $\Delta G(t)$ is the difference of the top two prediction scores,
% $
% \Delta G(t) := {\max}_1 G(t) - {\max}_2 G(t),
% $
% where $\max_i$ is the $i$-th largest value of the vector.}

To sample a subset of linear boundaries $\mathcal{Q}$ from $\mathcal{P}$, we sample a subset of reference images from $R$, and apply Equation~\eqref{eq:wb} on each of the sampled reference images to compute a linear boundary in $\mathcal{P}$.

%$G$ partitions $\Omega$ into a finite number of linear regions according to the active and inactive status of its hidden neurons; 
%and if the linear region containing $t^*$ intersects with $B$, 
%then the hyperplane computed in Equation~\eqref{eq:wb} is exactly a linear boundary $\mathbf{h}\in\mathcal{P}$.

By substituting the set of linear boundaries $\mathcal{P}$ in Equation~\eqref{eq:ideal_a} with the sampled set of linear boundaries $\mathcal{Q}$, we convert the optimization problem in Equation~\eqref{eq:ideal} into
\begin{subequations}
\label{eq:prac}
\begin{align}
	& \max_{P(x)\subseteq \mathcal{Q}} \left| P(x) \cap R \right| \label{eq:prac_a}\\
	& \text{s.t. } \left| P(x) \cap D(x) \right| \leq \delta \label{eq:prac_b},
\end{align}
\end{subequations}
where $\delta \geq 0$ is a relaxation parameter to ensure the above problem is feasible.
We discuss the effect of $\delta$ and how to compute $\delta$ as follows.

Recall that the linear boundaries in $\mathcal{P}$ partition $\Omega$ into a set of convex polytopes, and the images covered by the same convex polytope are predicted by $F$ as the same class~\cite{chu2018exact, montufar2014number}.  Since $\mathcal{Q}\subseteq \mathcal{P}$, the linear boundaries in $\mathcal{Q}$ also partition $\Omega$ into a set of convex polytopes.  However, due to the missing linear boundaries in $\mathcal{P} \setminus \mathcal{Q}$, some images covered by different neighbouring convex polytopes induced by $\mathcal{P}$ may be covered by the same convex polytope induced by $\mathcal{Q}$.
Therefore, for some convex polytopes induced by $\mathcal{Q}$, the images covered in the same convex polytope may not be predicted by $F$ as the same class.  In consequence, $\left| P(x) \cap D(x) \right|$ may be larger than zero.  Thus we use $\delta$ to keep the constraint in Equation~\eqref{eq:prac_b} valid, such that the problem in Equation~\eqref{eq:prac} is feasible.

Given an image $x\in\mathcal{X}$ to interpret, the minimum value of $\left| P(x) \cap D(x) \right|$, denoted by $\tau(x)$, is the number of images in $D(x)$ whose feature maps lie on the same side as $\psi(x)$ with respect to every linear boundary $\mathbf{h}\in\mathcal{Q}$.  We can easily compute $\tau(x)$ by enumeration.

Obviously, $P(x)=\mathcal{Q}$ is a feasible solution to the problem in Equation~\eqref{eq:prac} when $\delta \geq \tau(x)$.
We simply set $\delta=\tau(x)$, which yields spectacular interpretation performance in our experiments.

We prove that the problem in Equation~\ref{eq:prac} is a \textbf{submodular cost submodular cover (SCSC) problem}~\cite{Iyer2013SubmodularOW}.

%Then, we introduce a greedy algorithm~\cite{wolsey1982analysis} to solve it efficiently.

%We will illustrate how to compute $\delta$ later in this section.

%In this section, we will describe our algorithm that solves our optimization problem \eqref{optimize} and produces the output polytope $Q_{F}$.
%
%Recall that $\mathcal{P}$ is our candidate pool that consists of hyperplanes $H(x)=0$. For any subset $P$ of $\mathcal{P}$ and finite subset $X$ of $\mathcal{X}$, we can define
%\begin{equation}
%\begin{split}
%    S(X, P)~=~&\text{the set of all elements in $X$ \textbf{outside}}\\
%    &\text{of the convex polytope formed by $P$}
%\end{split}
%\end{equation}
%and $F(X, P)=|S(X, P)|$. Then, we have the following important property of $F$:

\begin{theorem}
\label{thm:scsc}
For a given input image $x\in\mathcal{X}$, the problem defined in Equation~\eqref{eq:prac} is a submodular cost submodular cover (SCSC) problem.
\end{theorem}

\proof See Appendix~\ref{apd:scsc} in \cite{lam2021finding}.

\begin{algorithm}[t]
\DontPrintSemicolon\small
\SetKwInput{KwInput}{Input}               
\SetKwInput{KwOutput}{Output} 
\KwInput{An input image $x\in\mathcal{X}$,  a trained CNN $F$, and a set of reference images $R\subseteq \mathcal{X}$.}
\KwOutput{A solution $P(x)$ to the SCSC problem in Equation~\eqref{eq:prac}.}

\textbf{Initialization}: $P(x)\leftarrow \emptyset$, $\delta \leftarrow \tau(x)$.

\While{$|P(x)\cap D(x)| > \delta$}{
	$\mathbf{h}\leftarrow\argmin\limits_{\mathbf{h}\in \mathcal{Q}\setminus P(x)} \frac{|P(x)\cap R| - |(P(x)\cup\{\mathbf{h}\})\cap R|}{|P(x)\cap D(x)| - |(P(x)\cup\{\mathbf{h}\})\cap D(x)|}$.\\

    	Add linear boundary $\mathbf{h}$ into the set of linear boundaries $P(x)$.\\
}

\Return $P(x)$.
\caption{A Greedy Algorithm}
\label{alg:greedy}
\end{algorithm}

\subsection{Solving the SCSC Problem}
An SCSC problem is NP-hard~\cite{crawford2019submodular},
thus we solve our problem in Equation~\eqref{eq:prac} by the simple greedy algorithm~\cite{wolsey1982analysis} summarized in Algorithm~\ref{alg:greedy}.

In each iteration of Algorithm~\ref{alg:greedy}, a boundary $\mathbf{h}\in\mathcal{Q}\setminus P(x)$ is selected to minimize 
\begin{equation}
\label{eq:policy}
	\frac{|P(x)\cap R| - |(P(x)\cup\{\mathbf{h}\})\cap R|}{|P(x)\cap D(x)| - |(P(x)\cup\{\mathbf{h}\})\cap D(x)|}.
\end{equation}

The numerator of Equation~\eqref{eq:policy} is the number of reference images that are originally covered by $P(x)$ but are removed from $P(x)$ after we add $\mathbf{h}$ into $P(x)$. 
Since our goal in Equation~\eqref{eq:prac_a} is to maximize the number of reference images covered by $P(x)$, we need to select an $\mathbf{h}$ that removes as few images in $R$ from $P(x)$ as possible.

Similarly, the denominator is the number of images in $D(x)$ that are originally covered by $P(x)$, but are removed from $P(x)$ after we add $\mathbf{h}$ into $P(x)$.
To quickly find a solution $P(x)$ that satisfies the constraint in Equation~\eqref{eq:prac_b}, we need to select an $\mathbf{h}$ that removes as many images in $D(x)$ from $P(x)$ as possible.

Following the above intuition, the greedy algorithm~\cite{wolsey1982analysis} in Algorithm~\ref{alg:greedy} adopts Equation~\eqref{eq:policy} to find an $\mathbf{h}\in\mathcal{Q}$ that minimizes the numerator and maximizes the denominator. In step 3 of Algorithm~\ref{alg:greedy}, when there are multiple linear boundaries that induce a numerator equal to zero, we pick the linear boundary $\mathbf{h}$ with the largest denominator to add into $P(x)$.
The denominator is positive before Algorithm~\ref{alg:greedy} converges, because we can always find a linear boundary to remove some images from $P(x)\cap D(x)$ until $|P(x)\cap D(x)| \leq \delta$.

The iteration continues until the constraint in Equation~\eqref{eq:prac_b} is satisfied.
Algorithm~\ref{alg:greedy} converges in at most $\left| \mathcal{Q} \right|$ iterations.

\section{Visualization and Similarity Ranking}
In this section, we first introduce how to visualize $P(x)$ as heat maps on the images covered by $P(x)$. Then, we discuss how to rank the images covered by $P(x)$ according to their semantic distances to the input image $x\in\mathcal{X}$.

\subsection{Visualizing $P(x)$ on Covered Images}
\label{sec:visual}

Since the heat maps for all the images covered by $P(x)$ can be produced in the same way, we take the input image $x$ as an example to introduce how to visualize $P(x)$.

To interpret why $F$ predicts the input image $x\in\mathcal{X}$ as $Class(x)$, we visualize $P(x)$ as a heat map that identifies the important regions in $x$ for the prediction. Since $F$ predicts every image covered by $P(x)$ as $Class(x)$, we interpret the decision logic of $F$ on $x$ by showing why $x$ is covered by $P(x)$. This is equivalent to showing why $\psi(x)$ is contained in $P(x)$. The key idea is to visualize $P(x)$ as a heat map to identify the important regions in $x$ that make $\psi(x)$ contained in $P(x)$. 

Recall that every linear boundary in $P(x)$ contributes to why $\psi(x)$ is contained in $P(x)$.
We first visualize every linear boundary in $P(x)$ as a heat map on $x$.
Then we produce the heat map of $P(x)$ by taking an average of the heat maps of all the linear boundaries.

Denote by $\langle W, t\rangle + b=0, t\in\Omega,$ the hyperplane of a linear boundary $\mathbf{h}$ in $P(x)$. 
We visualize $\mathbf{h}$ as a heat map on $x$ in a similar way as Grad-CAM~\cite{selvaraju2017grad}.
First, if $\langle W, \psi(x)\rangle + b <0$, then we set $W=-W$ and $b=-b$ to make sure $\langle W, \psi(x)\rangle + b >0$.
Second, we compute the \textbf{channel weight} $\alpha_k$ of the $k$-th channel of $\psi(x)$ by
\begin{equation}
	\alpha_k = \frac{1}{Z}\sum_i\sum_j W_{ij}^k,
\end{equation}
where $W_{ij}^k$ is the entry on the $i$-th row, the $j$-th column and the $k$-th channel in the tensor $W$, and $Z$ is the total number of entries $W_{ij}^k$ for every channel $k$.
Third, we compute a raw heat map by 
$raw=ReLU\left(\sum_k \alpha_k \psi^k(x) \right)$,
where $\psi^k(x)$ is the $k$-th channel of the feature map $\psi(x)$.  Last, we rescale and normalize the raw heat map to produce a heat map on $x$ in the same way as Grad-CAM. 
This heat map identifies the important regions in $x$ that show why $\mathbf{h}$ predicts $\psi(x)$ to be contained in $P(x)$.

Since $P(x)$ contains many linear boundaries, and every linear boundary contributes to why $\psi(x)$ is contained in $P(x)$, we compute the heat map for $P(x)$ as the average of the heat maps of all the linear boundaries.  Since the images covered by $P(x)$ are sharing exactly the same set of linear boundaries in $P(x)$, the channel weights computed from the $W$ of each linear boundary in $P(x)$ are exactly the same for all the images covered by $P(x)$. Therefore, we reuse the set of channel weights computed from the input image $x$ to produce heat maps for the other images covered by $P(x)$ in the same way as $x$.

As demonstrated later in Section~\ref{sec:qeud}, due to the high representativeness of $P(x)$, the heat maps produced by our method generalize well to unseen images, and they usually highlight conceptually similar parts of different images covered by the same $P(x)$.

\subsection{Ranking Similar Images}
\label{sec:ranking}

Now, let us rank the images covered by $P(x)$ according to their semantic distances to the input image $x\in\mathcal{X}$.

Recall that the linear boundaries in $P(x)$ identify the common decision logic of $F$ on all the images covered by $P(x)$. The common semantic of the covered images is captured by the \textbf{normal vectors} of the hyperplanes of these linear boundaries. Following this insight, we compute the semantic distance between the input image $x$ and another image $x'$ covered by $P(x)$ by first projecting their feature maps $\psi(x)$ and $\psi(x')$ onto the normal vectors, and then computing the L1-distance between the projected coordinates.

Denote by $\overrightarrow{W}_\mathbf{h}$ the normal vector of the hyperplane of a linear boundary $\mathbf{h}\in P(x)$.
We compute the semantic distance between $x$ and $x'$ by
\begin{equation}
	Dist(x, x') = \sum_{\mathbf{h}\in P(x)} \left| \langle \overrightarrow{W}_\mathbf{h}, \psi(x) \rangle - \langle \overrightarrow{W}_\mathbf{h}, \psi(x') \rangle \right|, 
\end{equation}
where $\langle \overrightarrow{W}_\mathbf{h}, \psi(x) \rangle$ and $\langle \overrightarrow{W}_\mathbf{h}, \psi(x') \rangle$ are the projected coordinates of $x$ and $x'$, respectively.  We rank the images covered by $P(x)$ according to their semantic distance to $x$ in ascending order, such that a more similar image with a smaller semantic distance is ranked higher in the list. 

As demonstrated in our case studies in Section~\ref{sec:cs}, showing similar images to the input image with conceptually similar heat maps makes our interpretations more convincing than showing the heat map for the input image alone.

%------------------------------------------------------------------------

\section{Experiments}
\label{sec:exp}

In this section, we present a systematic experimental study.  Limited by space, only selected results are reported in this section.  Please refer to the Appendix in \cite{lam2021finding} for more experimental results.

\subsection{Baseline Methods}
We conduct extensive experiments to evaluate the performance of our representative interpretation (RI) method, and compare with the state-of-the-art baselines including Automatic Concept-based Explanation (ACE)~\cite{Ghorbani2019TowardsAC} and three 
%\peter{(Removed the adjective "representative" here, might confuse the readers to think CAM-based methods are representative)}
CAM-based methods, Grad-CAM \cite{selvaraju2017grad}, Grad-CAM++ \cite{chattopadhay2018grad} and Score-CAM \cite{wang2020score}. We use the publicly available source code of the baseline methods.
The python notebook of RI can be found in \cite{ourcode}.
All baseline methods are compared using their optimal parameters.
For RI, the number of sampled linear boundaries in $\mathcal{Q}$ does not have much influence on the interpretation performance, thus we simply use $|\mathcal{Q}|=50$ for all the experiments. 

%For RI, the parameter $\epsilon$ of the binary search and the number of sampled linear boundaries in $\mathcal{Q}$ do not have much influence on the interpretation performance, we simply use $\epsilon=0.001$ and $|\mathcal{Q}|=50$ for all the experiments. 

We focus on evaluating the representativeness of interpretations, that is, how well an interpretation for an input image can be reused to interpret the predictions on new images that are similar to the input image~\cite{carvalho2019machine}.  Since the baseline methods cannot directly reuse their interpretations on new images, we propose the following extensions to reuse their interpretations.

The CAM-based methods use the channel weights computed from an input image $x\in\mathcal{X}$ to generate the interpretation (i.e., heat map) for $x$.  To reuse the interpretation generated by the CAM-based methods to interpret the prediction on a new image $x_{new}\in\mathcal{X}$, we reuse the channel weights computed from the input image $x$, and follow the same heat map producing steps of the CAM-based methods to generate the interpretation for $x_{new}$.  Recall that our method also reuses the set of channel weights computed from the input image $x$ to generate interpretations for new images covered by $P(x)$. It is fair to compare our method with the above extensions of the CAM-based methods.

ACE~\cite{Ghorbani2019TowardsAC} uses a set of input images in the same class to summarize important concepts. It clusters similar image segments and removes outlying image segments.  Every cluster of the similar image segments represents a concept.  To reuse these summarized concepts to interpret the prediction on a new image $x_{new}\in\mathcal{X}$, we use exactly the same pipeline as ACE to segment $x_{new}$ and remove outlying image segments. The remaining image segments are assigned to their nearest concepts, and are used to produce the interpretation on $x_{new}$ in the same way as ACE.

Another issue is that the baseline methods cannot identify the group of similar images that an interpretation can validly apply to.  To tackle this issue, following \cite{Ghorbani2019TowardsAC, zhang2018unreasonable}, we use Euclidean distance in the space of $\Omega$ to find the set of similar images whose feature maps are the closest to the feature map of the input image. 

\subsection{Datasets and Evaluation Metrics}

We use the following four public datasets to evaluate the performance of all the compared methods. 

The ASIRRA dataset~\cite{assira} contains two classes of images including `dog' and `cat'. The original dataset contains over 3 million images and a subset of them was made publicly available through Kaggle. The training data and the testing data contain 25,000 and 12,500 images, respectively.

The Gender Classification (GC) dataset~\cite{gender} contains two classes of images including `male' and `female'. The training data and the testing data contain 47,009 and 11,649 images, respectively.

The Retinal OCT Images (RO) dataset~\cite{oct} contains four classes of images, including one class of normal retina as `NORMAL', and three classes of retinal disease, `CNV', `DME' and `DRUSEN'.  The training data and the testing data contain 108,309 and 1,000 images, respectively.

The FOOD dataset~\cite{food} consists of two classes of images, `food' and `non-food'. 
The training data and the testing data contain 3,000 and 1,000 images, respectively. 

For the large datasets ASIRRA, GC and RO, we randomly sample 2,500 images from each class of the training data as the set of reference images $R$ to generate representative interpretations. For the small dataset FOOD, we use all the training images as the set of reference images.  We call a set of reference images a \textbf{reference dataset}. The ground truth labels of all reference images are not used to generate interpretations.

To evaluate how well an interpretation can be reused to interpret the predictions on unseen images, 
we randomly sample 1,000 images from the testing data of each dataset to produce an \textbf{unseen dataset}.

For each dataset, we train a VGG-19 model~\cite{Simonyan2015VeryDC} as the target model to interpret. Every model is trained using the model parameters pre-trained on ImageNet~\cite{deng2009imagenet} as initialization.

Next, we introduce how to quantitatively evaluate the representativeness of an interpretation. \peter{Denote by $S\subseteq \mathcal{X}$ the set of unseen evaluation images. For each interpretation method, we first generate some interpretations using the reference dataset, and then reuse these interpretations to identify the important regions in each image in $S$.} Then, we adopt the evaluation metrics in \cite{chattopadhay2018grad, ramaswamy2020ablation} to evaluate the representativeness of these interpretations. 
More specifically, we define 
\begin{equation}
\label{metrics_ad}
	\text{Average Drop (AD)} = \frac{1}{|S|}\sum_{e\in S}\frac{\max\left(0, Y_c(e)-Y_c(e') \right)}{Y_c(e)}
\end{equation}
and
\begin{equation}
\label{metrics_ai}
	\text{Average Increase (AI)} = \frac{1}{|S|}\sum_{e\in S} \mathbbm{1}_{Y_c(e) < Y_c(e')},
\end{equation}
where $Y_c(e)$ is the prediction score for class $c$ on an image $e\in S$, and $e'$ is a masked image produced by keeping 20\% of the most important pixels in $e$ according to the interpretation on $e$. The rest 80\% pixels in $e'$ are masked in black.

Since ACE uses image segments instead of heat maps as interpretations, it is hard to keep exactly 20\% pixels in a masked image.
Thus, we iteratively select the most important segments until more than 20\% pixels are kept.

A small AD and a large AI mean that the interpretation on the input image can be validly reused to accurately identify important regions \peter{on the images in $S$}. This further indicates a good representativeness performance of the interpretation.

\begin{figure}[t]
\centering
\includegraphics[width=82mm]{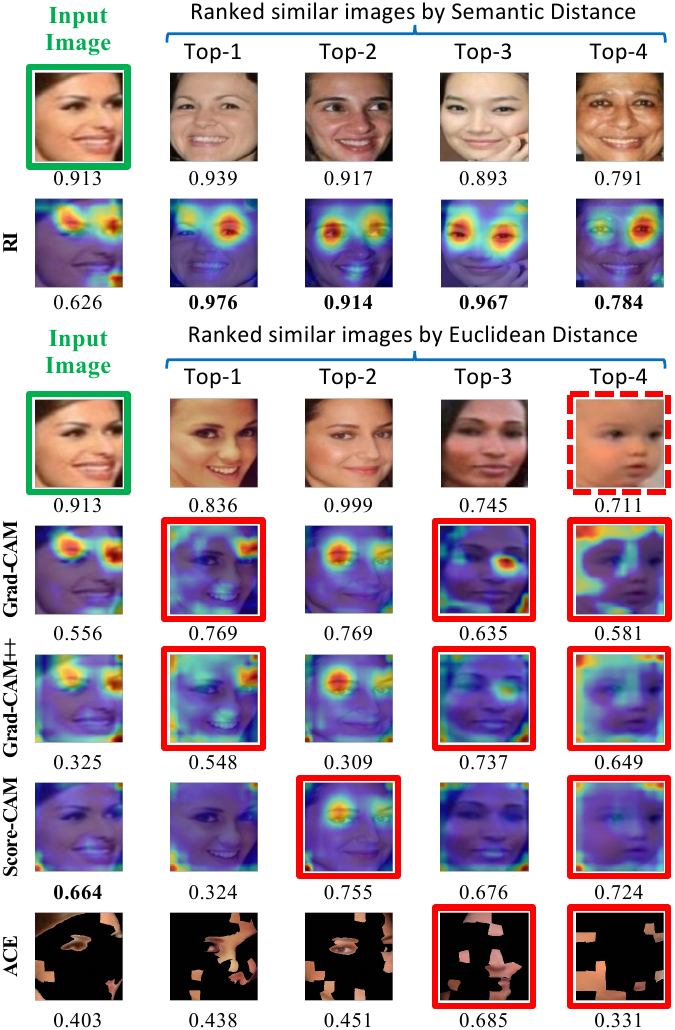}
\caption{A case study on the GC dataset. 
The first row shows the input image and the similar reference images retrieved by RI. 
The third row shows the same input image and the similar reference images retrieved by the baseline methods using Euclidean distance in $\Omega$. 
The other rows are the interpretations of different methods.
The numbers under the images in the first and the third rows are the prediction scores of the `female' class.
The numbers under the interpretations are the prediction scores of the `female' class on the masked images that are produced by keeping 20\% most important pixels.
A higher prediction score means a better interpretation.
}
\label{fig:case_female}
\end{figure}

\subsection{A Case Study}
\label{sec:cs}

%In this section, we use a case study to demonstrate the superior interpretation performance of RI over the other baseline methods.

In Figure~\ref{fig:case_female}, the first row shows an input image and the top-4 similar reference images in $R$ found by RI. The third row shows the same input image and the top-4 similar reference images found by the baseline methods using Euclidean distance. All images are predicted as `female' with high prediction scores.  The top-4 similar images found by RI are all adult females, which are conceptually very similar to the input image. Among the similar images found by the baselines, the fourth image highlighted by the red dashed box is a baby, which is very different from the input image. 

RI achieves a superior performance in finding similar images because the proposed co-clustering problem simultaneously finds the interpretation for the input image, as well as a set of images sharing the same interpretation.
Since all the images share the same interpretation, they are usually very similar in concepts.
The baseline methods are not designed to find similar images sharing the same interpretations. A straight-forward extension is to find the nearest neighbours using Euclidean distance in the space of $\Omega$, which, however, may not be effective in finding images with similar concepts.

To examine the representativeness of the interpretations of all the methods, we analyze whether an interpretation on the input image can be successfully reused on similar images to identify conceptually similar parts as the input image.  As shown in the second row of Figure~\ref{fig:case_female}, the interpretation computed by RI from the input image is highly representative because it consistently identifies the eyes and eyebrows of the females in the input image and the similar reference images.
These parts are significant features that distinguish a female from males in the GC dataset. Finding similar reference images with the same interpretation as the input image makes the interpretations produced by RI highly understandable and effective in practice.

The baseline methods cannot achieve representative interpretations.  Consider the images highlighted by the red boxes in Figure~\ref{fig:case_female}. The interpretations produced by the baseline methods cannot consistently identify the same conceptual parts for the input image and the similar reference images.
Especially, the interpretations produced by ACE are the most inconsistent due to the high sensitivity and randomness of image segmentation and clustering. Such inconsistency makes the interpretations ineffective.

Figure~\ref{fig:case_female} also shows the prediction score of `female' for every masked image 
produced by keeping 20\% of the most important pixels identified by an interpretation. 
A higher prediction score means a better interpretation quality.  Score-CAM achieves the best prediction score on the input image because it is designed to maximize the prediction score of the masked input image.
However, as shown in the second last row of Figure~\ref{fig:case_female}, the interpretation produced by Score-CAM cannot be effectively reused to achieve a high prediction score on the similar reference images.

RI achieves the second best prediction score on the input image and the best scores on all the similar reference images. This clearly indicates that the representative interpretation produced by RI is much more effective to be reused on similar images than the baselines. 

\subsection{Representativeness on Unseen Images}
\label{sec:qeud}

We quantitatively evaluate how well interpretations generated on a reference dataset can be reused to interpret predictions on unseen images.  For each method, we first generate a set of interpretations $V$ on the reference dataset. Then, we use metrics AD and AI to evaluate the representativeness of  $V$ on the unseen dataset.

To generate $V$ for each of RI, Grad-CAM, Grad-CAM++ and Score-CAM, we first apply $k$-means~\cite{lloyd1982least} on the reference dataset to find $|V|$ clusters in the space $\Omega$. 
Then, we find the reference images that are the closest to each of the cluster centers, and use them as the input images to produce the interpretations in $V$.  Since ACE requires a set of reference images as the input to produce one interpretation, we first find the $|V|$ sets of reference images that form the $|V|$ clusters produced by the $k$-means algorithm, and then use each set of reference images as an input to ACE to produce an interpretation in $V$.

The number of interpretations in $V$ is set to
$|V|=10\times\text{Number of classes}$ for all methods, 
that is, $|V|=40$ for RO, and $|V|=20$ for the other datasets.

\peter{To evaluate the representativeness of the group of interpretations $V$ produced by RI, for each image $e\in S$, we 
first find the polytope in $v\in V$ that covers $e$, and then we use $v$ to interpret $e$. If an unseen image is not covered by the convex polytope of any interpretation in $V$, 
we find its nearest cluster center in the space $\Omega$, and use the polytope it generates to interpret $e$.}

Since the baseline methods cannot find the set of unseen images that an interpretation can apply to,
we assign each unseen image in the unseen dataset to its nearest cluster center in the space $\Omega$. 
%\peter{(Removed "In this way ... AD and AI.")}
% In this way, we partition the unseen dataset into $|V|$ sets of unseen images. We use each set as the $S_v$ for the corresponding interpretation $v\in V$ to compute the metrics AD and AI.

Table~\ref{tab:unseen} shows the Average Drop (AD) and Average Increase (AI) of all the methods.  We run 5 independent experiments for each method, and report the mean and standard deviation of mAD and mAI.

RI achieves the best mAD performance on all the datasets and the best mAI performance on most of the datasets. This clearly domonstrates the superior performance of RI in finding representative interpretations that reveals the common decision logic on the predictions of a large number of similar unseen images.

\begin{table}[t]
\begin{center}
\small
\scalebox{0.85}{
\begin{tabular}{c|c|c|c}
\toprule
%\multirow{2}{*}{Datasets} & \multirow{2}{*}{Methods} & \multicolumn{2}{c}{Mean AD (\%)} & \multicolumn{2}{c}{Mean AI (\%)} \\
Datasets & Methods & mAD (\%) & mAI (\%) \\

\midrule

\multirow{5}{*}{ASIRRA} 
& RI & \textbf{2.00} $\pm$ 0.06 & 30.72 $\pm$ 0.85 \\
& Grad-CAM & 2.30 $\pm$ 0.09 & \textbf{32.90} $\pm$ 0.39  \\
& Grad-CAM++ & 3.07 $\pm$ 0.02 & 29.12 $\pm$ 0.42  \\
& Score-CAM & 3.72 $\pm$ 0.09 & 28.10 $\pm$ 0.28 \\
& ACE & 36.27 $\pm$ 0.78 & 1.36 $\pm$ 0.2  \\

\midrule

\multirow{5}{*}{GC} 
& RI & \textbf{12.70} $\pm$ 0.32 & \textbf{6.12} $\pm$ 0.19 \\
& Grad-CAM & 23.01 $\pm$ 0.50 & 4.40 $\pm$ 0.33  \\
& Grad-CAM++ & 29.53 $\pm$ 0.36 & 3.46 $\pm$ 0.40  \\
& Score-CAM & 23.88 $\pm$ 0.37 & 3.30 $\pm$ 0.38 \\
& ACE & 46.37 $\pm$ 0.93 & 2.26 $\pm$ 0.12  \\

\midrule

\multirow{5}{*}{RO} 
& RI & \textbf{12.57} $\pm$ 0.17 & \textbf{27.54} $\pm$ 0.50 \\
& Grad-CAM & 35.34 $\pm$ 0.78 & 22.02 $\pm$ 0.33  \\
& Grad-CAM++ & 21.04 $\pm$ 0.33 & 23.24 $\pm$ 0.59  \\
& Score-CAM & 14.12 $\pm$ 1.39 & 25.60 $\pm$ 0.34 \\
& ACE & 65.36 $\pm$ 0.91 & 4.44 $\pm$ 0.39  \\

\midrule

\multirow{5}{*}{FOOD} 
& RI & \textbf{8.08} $\pm$ 0.33 & \textbf{26.66} $\pm$ 0.48 \\
& Grad-CAM & 9.22 $\pm$ 0.64 & 23.34 $\pm$ 1.12  \\
& Grad-CAM++ & 9.04 $\pm$ 0.27 & 22.14 $\pm$ 1.34  \\
& Score-CAM & 10.44 $\pm$ 1.84 & 19.56 $\pm$ 0.89 \\
& ACE & 39.88 $\pm$ 2.13 & 24.80 $\pm$ 0.76  \\

\bottomrule
\end{tabular}
}
\end{center}
\caption{The mean Average Drop (mAD) and the mean Average Increase (mAI) of all the methods.}
\label{tab:unseen}
\end{table}

%------------------------------------------------------------------------

\section{Conclusion}

In this paper, we develop a novel unsupervised method to produce representative interpretation for many similar images. We formulate a co-clustering problem to simultaneously find the largest group of similar images and the common decision logic of a CNN. To make our interpretations more understandable, we visualize the common decision logic as a heat map to highlight important image regions, and also rank the similar images according to their semantic distances to the input image. Our extensive experiments demonstrate the superior representativeness and quality of the interpretations produced by our method.

%------------------------------------------------------------------------

\section{Acknowledgement}
Lingyang Chu's research is supported in part by the start-up grant provided by the Department of Computing and Software of McMaster University. 
All opinions, findings, conclusions and recommendations in this paper are those of the authors and do not necessarily reflect the views of the funding agencies.

{\small
\bibliographystyle{ieee_fullname}
\bibliography{egbib}
}

\appendix

\clearpage
\appendix

\begin{center}
	{\Large \bf APPENDIX}
\end{center}

In this appendix, we provide the proof for Theorem~\ref{thm:scsc} in Section~\ref{apd:scsc}, and provide more experiment results to demonstrate the outstanding interpretation performance of RI in Sections~\ref{sec:cs2}-\ref{sec:extra}.

\section{The Proof of Theorem~\ref{thm:scsc}}
\label{apd:scsc}

We first rewrite Equation~\eqref{eq:prac} as
\begin{subequations}
\label{eq:cvt}
\begin{align}
	& \min_{P(x)\subseteq \mathcal{Q}} \left| R \right| - \left| P(x) \cap R \right| \\
	& \text{s.t. } \left| D(x) \right| - \left| P(x) \cap D(x) \right| \geq \left| D(x) \right| - \delta, 
\end{align}
\end{subequations}
where $\left| R \right|$ and $\left| D(x) \right|$ are the constant cardinalities of $R$ and $D(x)$, respectively.

Then, we prove the problem of Equation~\eqref{eq:cvt} is an SCSC problem~\cite{Iyer2013SubmodularOW} by showing 
\begin{equation}
	\left| R \right| - \left| P(x) \cap R \right|
\end{equation}
and
\begin{equation}
	\left| D(x) \right| - \left| P(x) \cap D(x) \right|
\end{equation}
are submodular functions with respect to $P(x)\subseteq\mathcal{Q}$.

%We only need to prove $\left| D \right| - \left| P(x) \cap D \right|$ is a submodular function with respect to $P(x)$, because 

Denote by $P=P(x)$, and by $f(P) = \left| R \right| - \left| P \cap R \right|$.
We prove $f(P)$ is a submodular function with respect to $P\subseteq\mathcal{Q}$ by showing that, 
for any two sets of linear boundaries, denoted by $P\subseteq \mathcal{Q}$ and $T\subseteq \mathcal{Q}$, if $P\subseteq T$, then 
\begin{equation}
\label{eq:sbm}
	f(P\cup\{\mathbf{h}\})-f(P) \geq f(T\cup\{\mathbf{h}\})-f(T)
\end{equation}
holds for any linear boundary $\mathbf{h}\in\mathcal{Q}\setminus T$.

Recall that the linear boundaries in $P$ defines a convex polytope, and $\left| P \cap R \right|$ is the number of images in $R$ that are covered by $P$.
Thus, $f(P)=\left| R \right| - \left| P \cap R \right|$ is the number of images in $R$ that are not covered by $P$. 

If we add a new linear boundary $\mathbf{h}\in\mathcal{Q}\setminus T$ to $P$, some images covered by $P$ may not be covered by the new convex polytope defined by $P\cup\{\mathbf{h}\}$.
We say these images are removed by $\mathbf{h}$ from $P$.
%be separated by $\mathbf{h}$ from the convex polytope defined by $P\cup\{\mathbf{h}\}$.
%We say these images are separated from $P\cup\{\mathbf{h}\}$.

The left side of Equation~\eqref{eq:sbm} is exactly the number of the images removed by $\mathbf{h}$ from $P$. 
Similarly, the right side of Equation~\eqref{eq:sbm} is the number of the images removed by $\mathbf{h}$ from $T$.

Since $P\subseteq T$, the set of images covered by $P$ contains the set of images covered by $T$.
Therefore, the number of images removed by $\mathbf{h}$ from $P$ will be no smaller than the number of images removed by $\mathbf{h}$ from $T$.
As a result, Equation~\eqref{eq:sbm} holds, which means $\left| R \right| - \left| P(x) \cap R \right|$ is a submodular function with respect to $P(x)$.

We can prove $\left| D(x) \right| - \left| P(x) \cap D(x) \right|$ is a submodular function with respect to $P(x)$ in the same way.
This concludes the theorem.

\begin{figure}
\centering
\includegraphics[width=80mm]{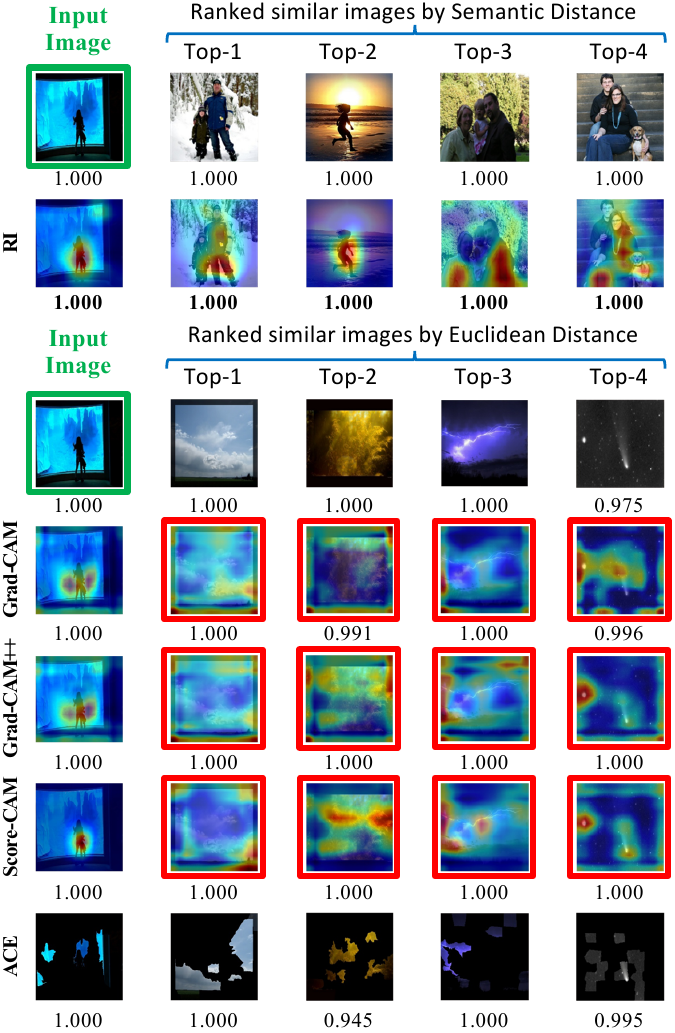}
\caption{A case study on the FOOD dataset. 
The first row shows the input image and the similar reference images found by RI. 
The third row shows the same input image and the similar reference images found by the baseline methods. 
The other rows are the interpretations of all the methods.
The numbers under the images in the first and third row are the prediction scores of the `non-food' class.
The numbers under the interpretations are the prediction scores of `non-food' on the masked images produced by keeping 20\% most important pixels.
A higher prediction score means a better interpretation.
}
\label{fig:food}
\end{figure}

\section{Case Study}
\label{sec:cs2}

We present a case study on the FOOD dataset in Figure~\ref{fig:food}.
The experiment setting is the same as the case study discussed in Section~\ref{sec:cs}.

As shown in the first row of Figure~\ref{fig:food}, the input image and the similar reference images found by RI are all predicted as `non-food' with high prediction scores.

All these images contain human, which indicates the common decision logic to predict these images as `non-food' is that they contain human bodies instead of food. 

We can see from the second row of Figure~\ref{fig:food} that this common decision logic is accurately identified by RI. 
The interpretation computed by RI from the input image correctly identifies the human bodies in the input image and the similar reference images.

These results demonstrate the superior performance of RI in finding representative interpretations that reveal the common decision logic on similar images.

The third row of Figure~\ref{fig:food} shows the same input image as the first row, and the similar reference images found by the baseline methods. 
These images are perceived as conceptually similar because they all contain large areas of natural scene, thus the common decision logic to predict them as `non-food' should be the natural scene.

However, we can see from the 4th, 5th and 6th rows that the interpretations produced by Grad-CAM, Grad-CAM++ and Score-CAM all identify the human bodies instead of the natural scene in the input image, but the interpretations on the similar reference images all identify natural scene.

The inconsistency between the interpretations on the input image and the similar reference images demonstrate that Grad-CAM, Grad-CAM++ and Score-CAM cannot produce good representative interpretations to reveal the common decision logic of a CNN in making predictions on a large number of similar images.

We can also see from the last row of Figure~\ref{fig:food} that ACE did a fairly good job in consistently identifying the image patches of natural scene in the input image and the reference images.
However,
as demonstrated by the quantitative experimental results in Section~\ref{sec:qeud} and Appendix~\ref{sec:ref}, the interpretation quality of ACE is significantly lower than RI due to the high sensitivity of image segmentation and clustering.

\section{Representativeness on Reference Images}
\label{sec:ref}

In this section, we use the reference dataset to quantitatively evaluate how well can an interpretation generated on an input image be reused to interpret predictions made on similar reference images.

\newcommand{\figurewidth}{40mm}
\begin{figure}[t]
\centering
%===================================================
\subfigure[mAD on ASIRRA]{\includegraphics[width=\figurewidth]{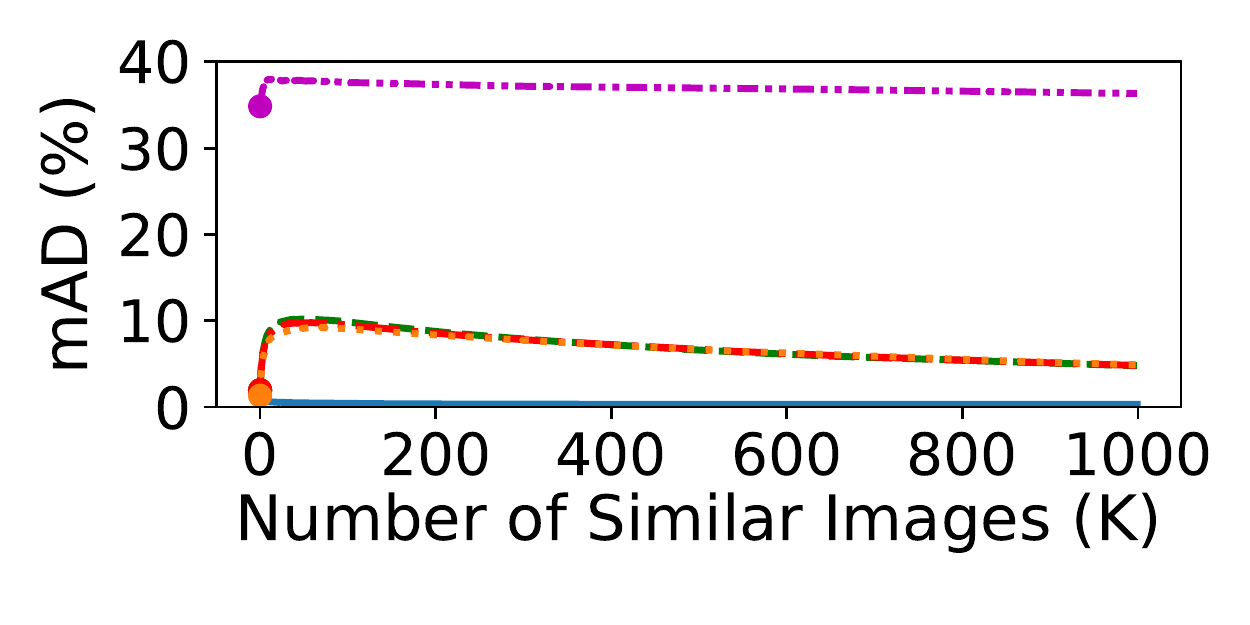}}
\subfigure[mAI on ASIRRA]{\includegraphics[width=\figurewidth]{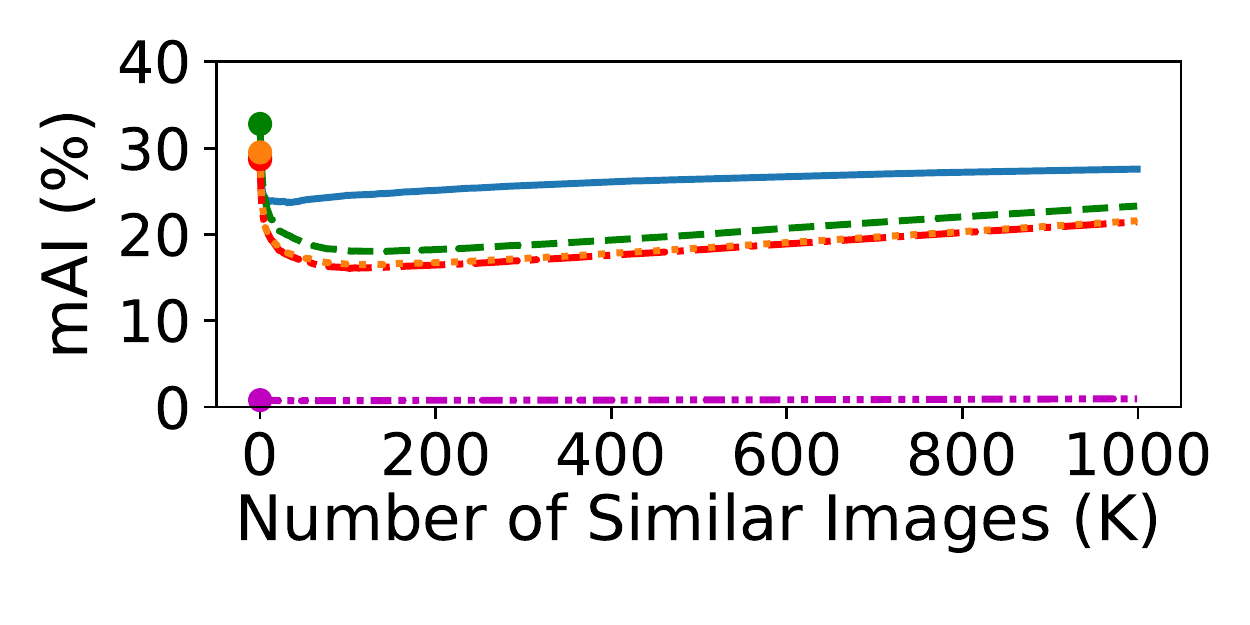}}
%===================================================
\subfigure[mAD on GC]{\includegraphics[width=\figurewidth]{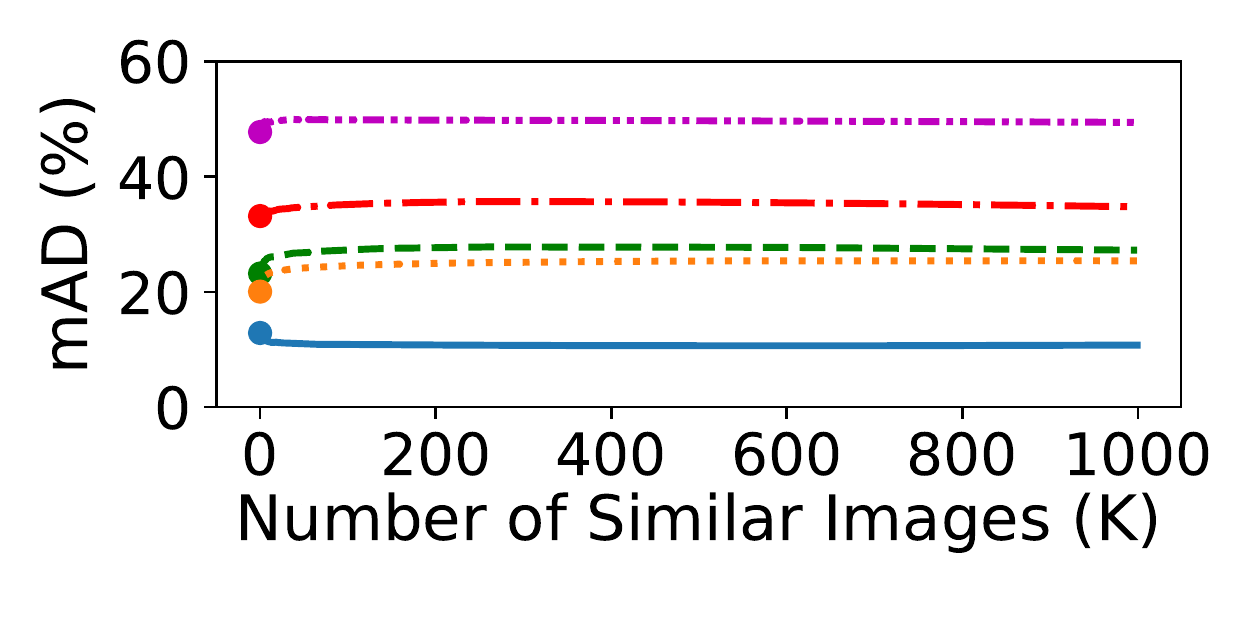}}
\subfigure[mAI on GC]{\includegraphics[width=\figurewidth]{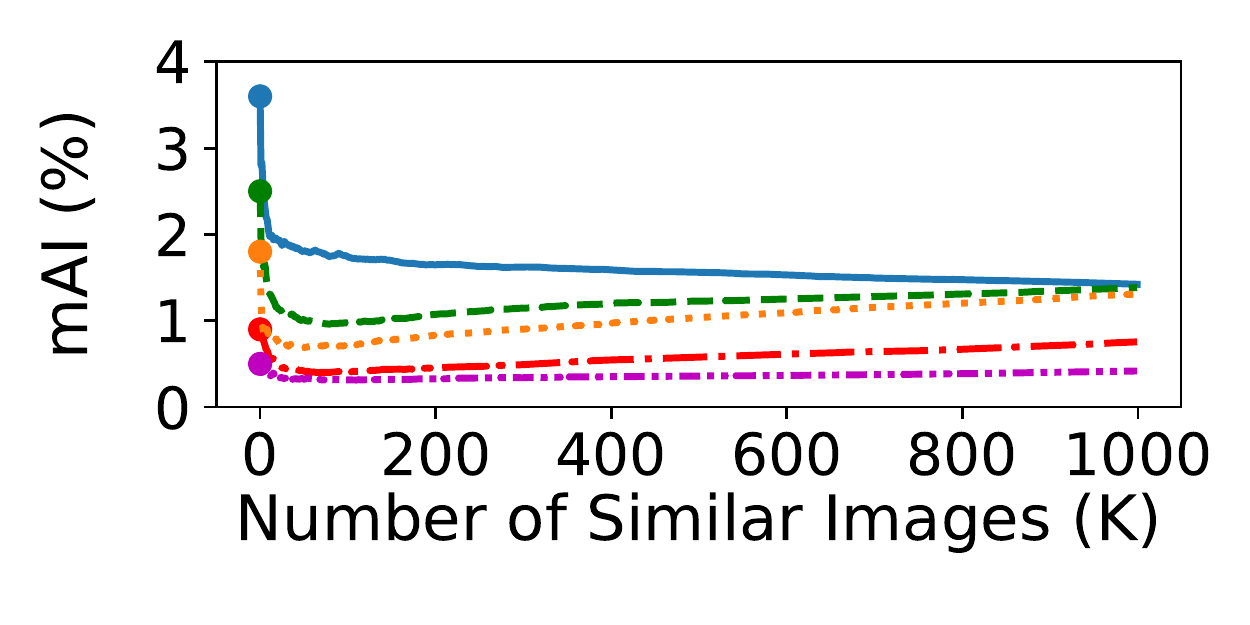}}
%===================================================
\subfigure[mAD on RO]{\includegraphics[width=\figurewidth]{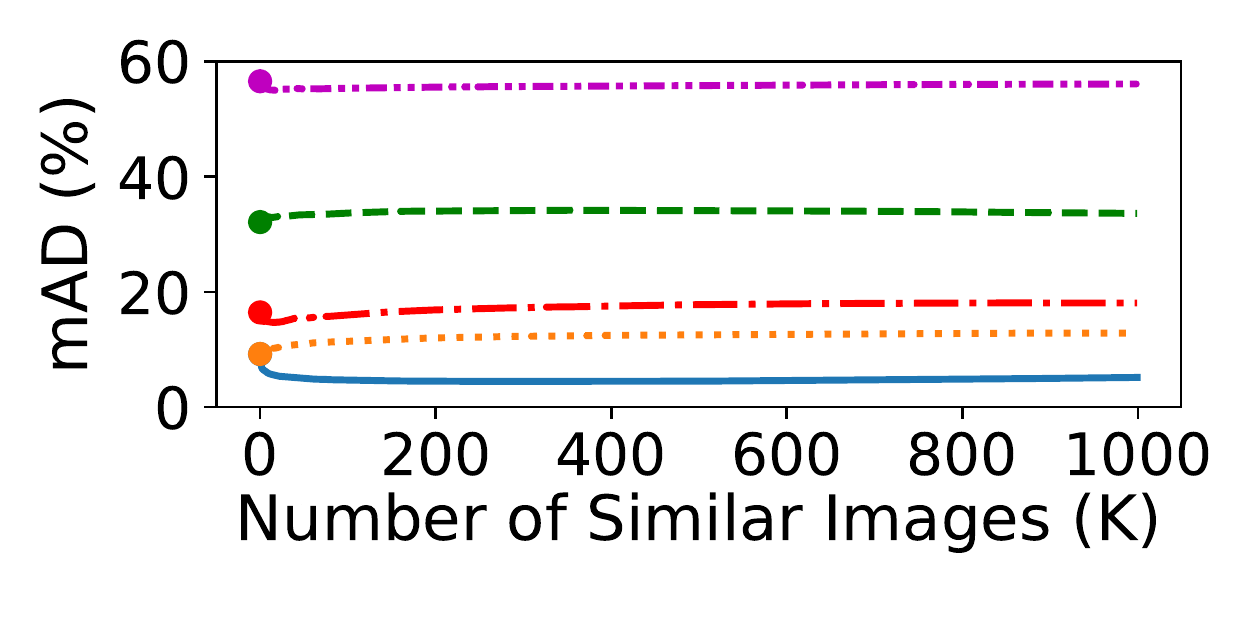}}
\subfigure[mAI on RO]{\includegraphics[width=\figurewidth]{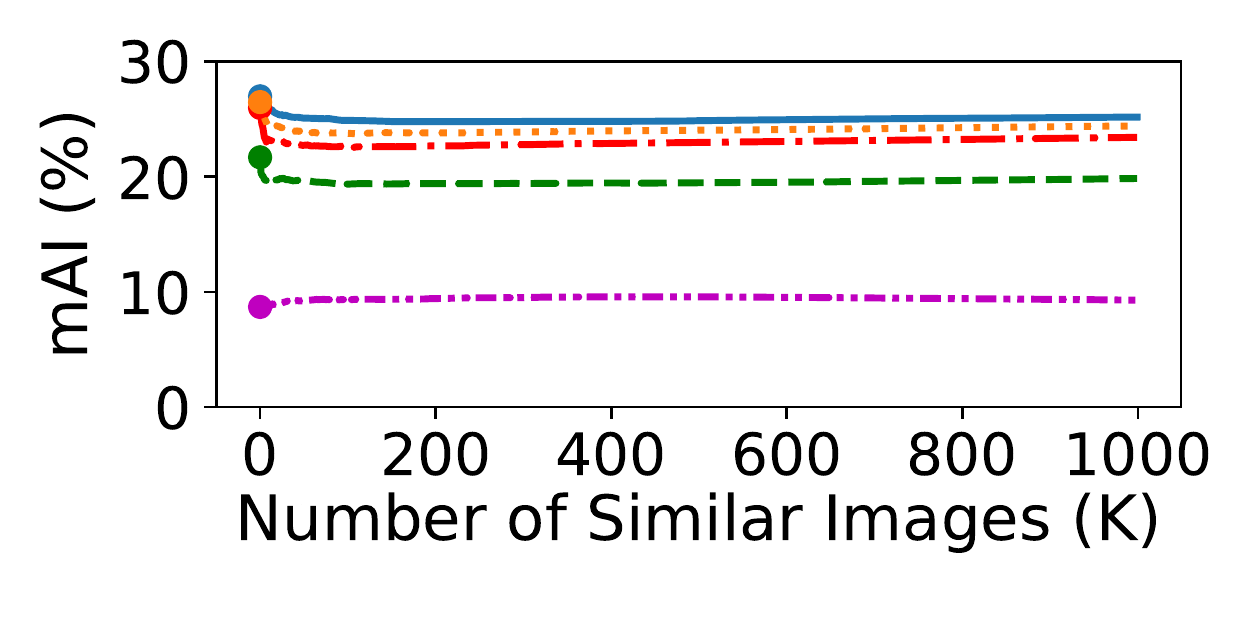}}
%===================================================
\subfigure[mAD on FOOD]{\includegraphics[width=\figurewidth]{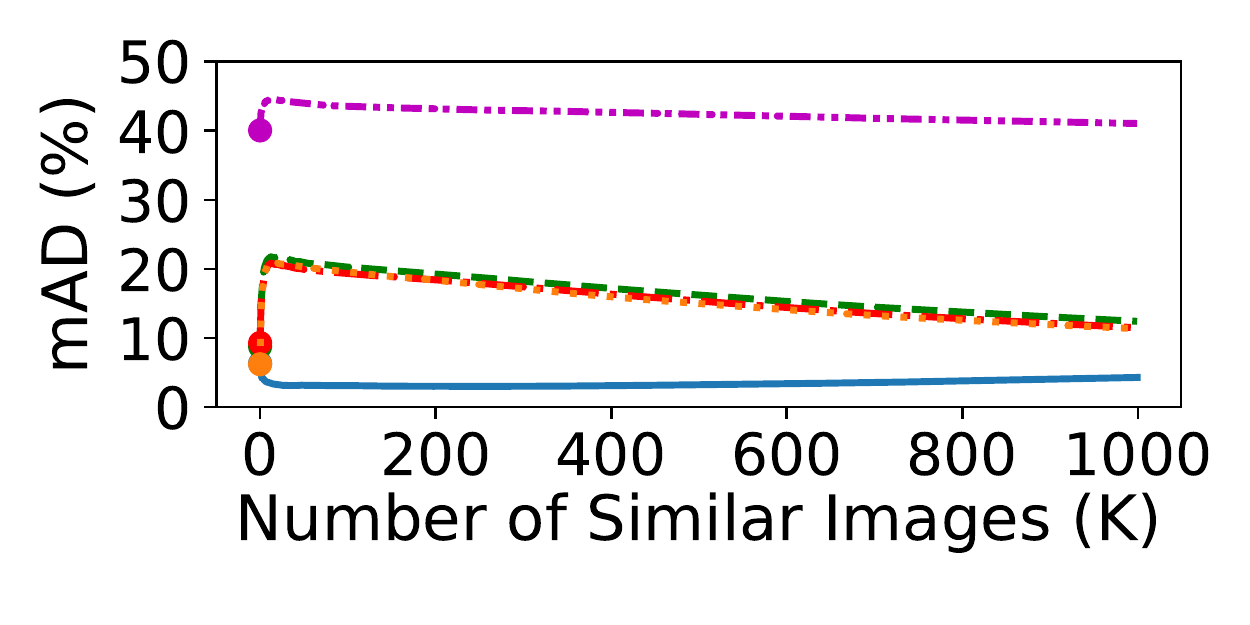}}
\subfigure[mAI on FOOD]{\includegraphics[width=\figurewidth]{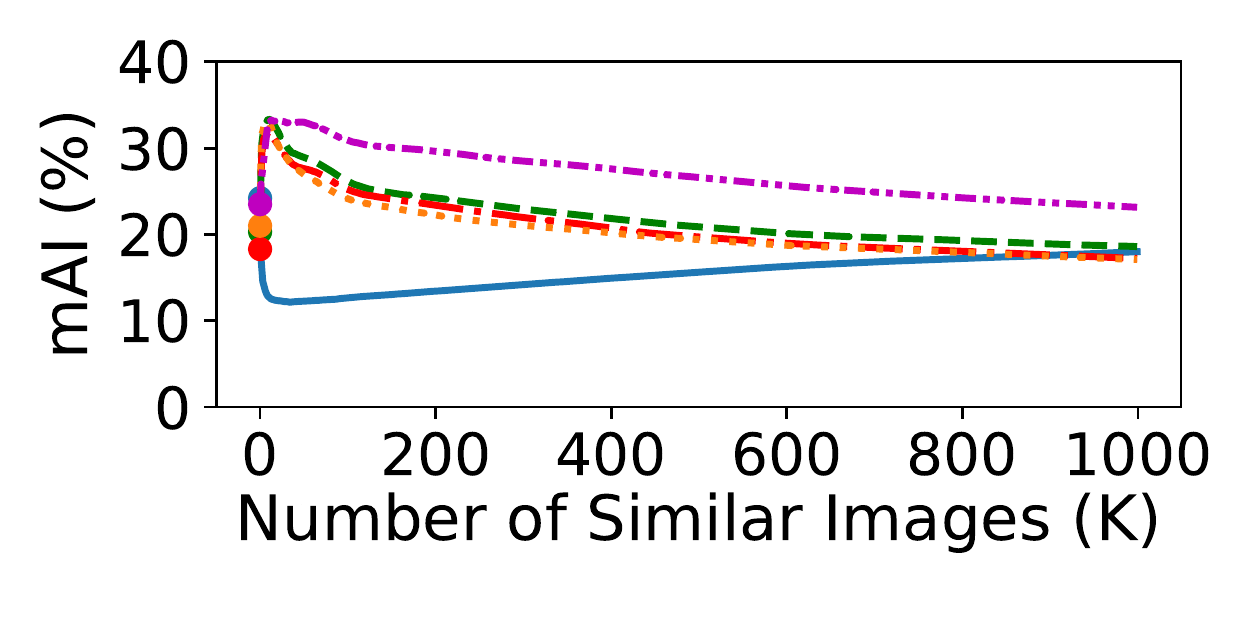}}
%===================================================
\includegraphics[width=0.45\textwidth]{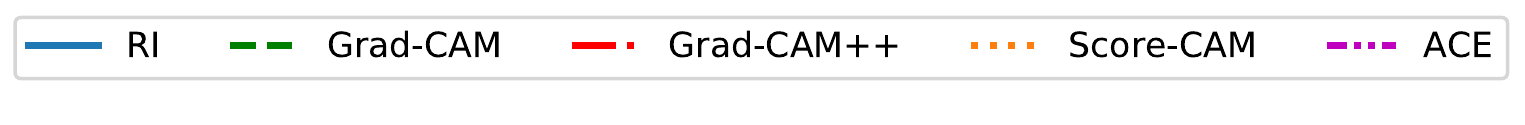}
\caption{The mean Average Drop (mAD) and mean Average Increase (mAI) performance of all the methods. 
}
\label{fig:exp_ref}
\end{figure}

\begin{table*}
\footnotesize
\begin{center}
\begin{tabular}{cc|cc|cc|cc}
\toprule

\multicolumn{2}{c|}{Datasets} & ACU-GT of RI (\%) & ACU-GT of $F$ (\%) & ACU-MD of RI (\%) & ACU-MD of $F$ (\%) & CVR of RI (\%) & CVR of $F$ (\%) \\

\midrule

\multirow{2}{*}{ASIRRA} & Ref. & $99.85\pm0.04$ & 100.00 & $99.85\pm0.04$  & 100.00 & $99.83\pm0.05$ & 100.00\\
& Unseen & $98.38\pm0.12$ & 98.80 & $99.28\pm0.08$ & 100.00 & $99.84\pm0.14$ & 100.00\\

\midrule

\multirow{2}{*}{GC} & Ref. & $96.25\pm0.09$ & 95.34 & $99.63\pm0.05$ & 100.00 & $97.95\pm0.14$ & 100.00\\
& Unseen & $94.95\pm0.13$ & 94.00 & $98.98\pm0.20$ & 100.00 & $98.18\pm0.39$ & 100.00\\

\midrule

\multirow{2}{*}{RO} & Ref. & $98.63\pm0.08$ & 99.58 & $98.98\pm0.08$ & 100.00 & $99.69\pm0.05$ & 100.00\\
& Unseen & $98.14\pm0.17$ & 97.50 & $98.44\pm0.10$ & 100.00 & $99.88\pm0.15$ & 100.00\\

\midrule

\multirow{2}{*}{FOOD} & Ref. & $99.84\pm0.06$ & 100.00 & $99.84\pm0.06$ & 100.00 & $99.45\pm0.11$ & 100.00\\
& Unseen & $97.87\pm0.13$ & 98.50 & $98.54\pm0.10$ & 100.00 & $99.68\pm0.21$ & 100.00\\

\bottomrule
\end{tabular}
\end{center}
\caption{
The ACU-GT, ACU-MD and CVR performance of RI and the CNN model $F$. 
We run RI for 5 independent times and compute the mean and standard deviation for each of ACU-GT, ACU-MD and CVR.
The ACU-GT of $F$ is the prediction accuracy of the CNN model $F$ with respect to the ground truth labels.
The ACU-MD of $F$ is always 100\% because the prediction results of $F$ are used as the ground truth to compute ACU-MD.
The CVR of $F$ is always 100\% because $F$ is applicable to predicting the labels of all the images.
Ref. and Unseen represents the reference dataset and the unseen dataset, respectively.
We set $|V|=40$ for RO, and $|V|=20$ for the other datasets.
%The prediction accuracy and coverage of the decision regions extracted by RI. Avg. ACU-GT is the average accuracy of decision regions that is computed using the ground truth labels of the datasets. Avg. ACU-MD is the average accuracy of the decision regions that is computed by using the labels predicted by the CNN as ground truth. CVR is the coverage of the convex polytopes of the decision regions.
}
\label{tab:cov}
\end{table*}

We randomly sample 1,000 images from the reference dataset as the input images to generate 1,000 interpretations for each method. For each interpretation, we use AD \eqref{metrics_ad} and AI \eqref{metrics_ai} again for evaluation,\peter{ except this time we take $S$ to be the 1,000 most similar reference images.}

For each of RI, Grad-CAM, Grad-CAM++ and Score-CAM, we use every input image to generate an interpretation.
Since ACE requires a set of images to produce one interpretation, we first use an input image to find 49 nearest reference images to the input image in the space of $\Omega$. Then, we use the 50 images including the input image and the 49 nearest images as the input to ACE to generate one interpretation for the input image. 

For each of the baseline methods, 
to evaluate the representativeness of an interpretation computed from an input image $x$, 
we first find the top-$K$ nearest reference images to $x$ in the space of $\Omega$ using Euclidean distance, and then use these images as the set of similar images $S$ to compute the AD and AI of the interpretation.

For RI, we use the set of top-$K$ similar images ranked by semantic distance to compute the AD and AI of each interpretation.

Figure~\ref{fig:exp_ref} shows the mean Average Drop (mAD) and the mean Average Increase (mAI) of the 1,000 interpretations generated by each of the methods for different values of $K$.

Here, $K=1$ means the set of similar images $S$ contains only the input image $x$, because $x$ is the most similar to itself.
In this case, the mAD and mAI performance are evaluated on the input images only.

We can see that RI achieves the best mAD performance on all the datasets, and it also achieves the best mAI performance on most of the datasets.
These results demonstrate the superior performance of RI in producing representative interpretations.

The mAI performance of RI is worse than the other methods in Figure~\ref{fig:exp_ref}(h), because a large proportion of the similar images found by RI on the FOOD dataset has a high prediction score close to 100\%.
Thus, it is very difficult to further increase the score by masking the images based on interpretations.

Actually, due to the high quality of the representative interpretations produced by RI, most of the masked images produced by RI only have a slight drop of prediction scores. Therefore, as shown in Figure~\ref{fig:exp_ref}(g), RI still achieves a much better mAD performance than all the baseline methods on the FOOD dataset.

We can also see in Figures~\ref{fig:exp_ref}(a), \ref{fig:exp_ref}(e) and \ref{fig:exp_ref}(g) that the mAD of Score-CAM is slightly better than RI when $K=1$. This is because Score-CAM focuses on maximizing the prediction scores of the masked input image, thus it achieves a better mAD on the input image.
However, RI focuses on finding the most representative interpretation for both the input image and a large number of similar images.
Therefore, 
when $K=1$, RI achieves a comparable performance with Score-CAM on the input images; 
and when $K>1$, RI achieves a much better performance than Score-CAM on the similar images.

\section{Prediction Accuracy and Coverage of the Decision Regions Produced by RI}
\label{sec:cov}

In this section, we evaluate the quality of the interpretations of RI by analyzing the prediction accuracy and the coverage of the decision regions produced by RI.

Recall that each representative interpretation produced by RI for an input image $x\in\mathcal{X}$ is a convex polytope $P(x)$, and $P(x)$ induces a decision region that predicts all the images it covers as $Class(x)$.

%The quality of the decision region induced by $P(x)$ is closely related to the quality of the corresponding interpretation.

If the predictions made by a decision region have a high accuracy, then the corresponding interpretation will be closer to the real decision logic of the CNN $F$.
If the convex polytope of the decision region covers a lot of images, then the interpretation is representative.

Based on the above insight, we design the following experiments to analyze the prediction accuracy and the coverage of the decision regions produced by RI.

For each dataset, we first follow the steps in Section~\ref{sec:qeud} to generate a set $V$ of interpretations using RI.
Each interpretation $v\in V$ corresponds to a convex polytope that induces a decision region.
This gives us a number of $|V|$ decision regions in total.

As illustrated in Section~\ref{sec:qeud}, we set $|V|=40$ for RO, and $|V|=20$ for the other datasets.

For each decision region produced by RI from an input image $x$, we predict all the images covered by the corresponding convex polytope as $Class(x)$.

If an image $x'\in\mathcal{X}$ is covered by multiple convex polytopes, we predict the label of $x'$ by the decision region induced by the convex polytope that covers the largest number of reference images.

We evaluate the accuracy of the predictions made by the decision regions of RI by the following two types of prediction accuracies.

The first one, denoted by ACU-GT, is the prediction accuracy computed using the ground truth labels of the images.
A higher ACU-GT means the decision regions work better in accurately predicting the ground truth labels of the covered images, which further indicates the corresponding interpretations are likely to capture some useful patterns for making accurate predictions.

The second one, denoted by ACU-MD, is the prediction accuracy computed by treating the labels predicted by the CNN model $F$ as ground truth.
A higher ACU-MD means the predictions made by the decision regions align better with the predictions made by $F$, which further indicates the corresponding interpretations are closer to the decision logic of $F$.

%For each of the $|V|$ decision regions produced by RI, we compute an ACU-GT and an ACU-MD on the images covered by the decision region.

We also evaluate the coverage (CVR) of the decision regions by the proportion of images that are covered by at least one of the $|V|$ decision regions.
A larger CVR means a higher representativeness of the interpretations.

We run RI for 5 independent times and compute the mean and standard deviation for each of ACU-GT, ACU-MD and CVR.

Table~\ref{tab:cov} shows the ACU-GT, ACU-MD and CVR performance of RI and the CNN model $F$ on the reference datasets and unseen datasets of ASIRRA, GC, RO and FOOD.

Since the interpretations produced by the baseline methods cannot be used as classifiers to make predictions on images, we cannot report their ACU-GT, ACU-MD and CVR performance.

As shown in Table~\ref{tab:cov}, the decision regions produced by RI achieve very high ACU-GT on both the reference dataset and the unseen dataset. This means the decision regions capture some useful representative patterns to make accurate predictions.

The ACU-GT of RI sometimes outperforms the original CNN model $F$.
This is because the ACU-GT of RI is computed on the covered images, but the ACU-GT of $F$ is computed on the complete set of images, which may contain more misclassified images.

The decision regions also achieve very high ACU-MD on all the datasets. This indicates that the predictions made by the decision regions align very well with the CNN model $F$. Thus, the corresponding interpretations are closer to the decision logic of $F$.

The CVR of RI is also very large. With 40 interpretations for RO and 20 interpretations for the other datasets, the proportion of images covered by the corresponding decision regions is more than 97\% on all the datasets.

Recall that an interpretation of RI generally applies to all the image covered by the corresponding decision region, a large CVR demonstrates the outstanding representativeness of the interpretations produced by RI.

\begin{figure}[t]
\centering
%===================================================
\subfigure[Reference dataset of ASIRRA]{\includegraphics[width=0.22\textwidth]{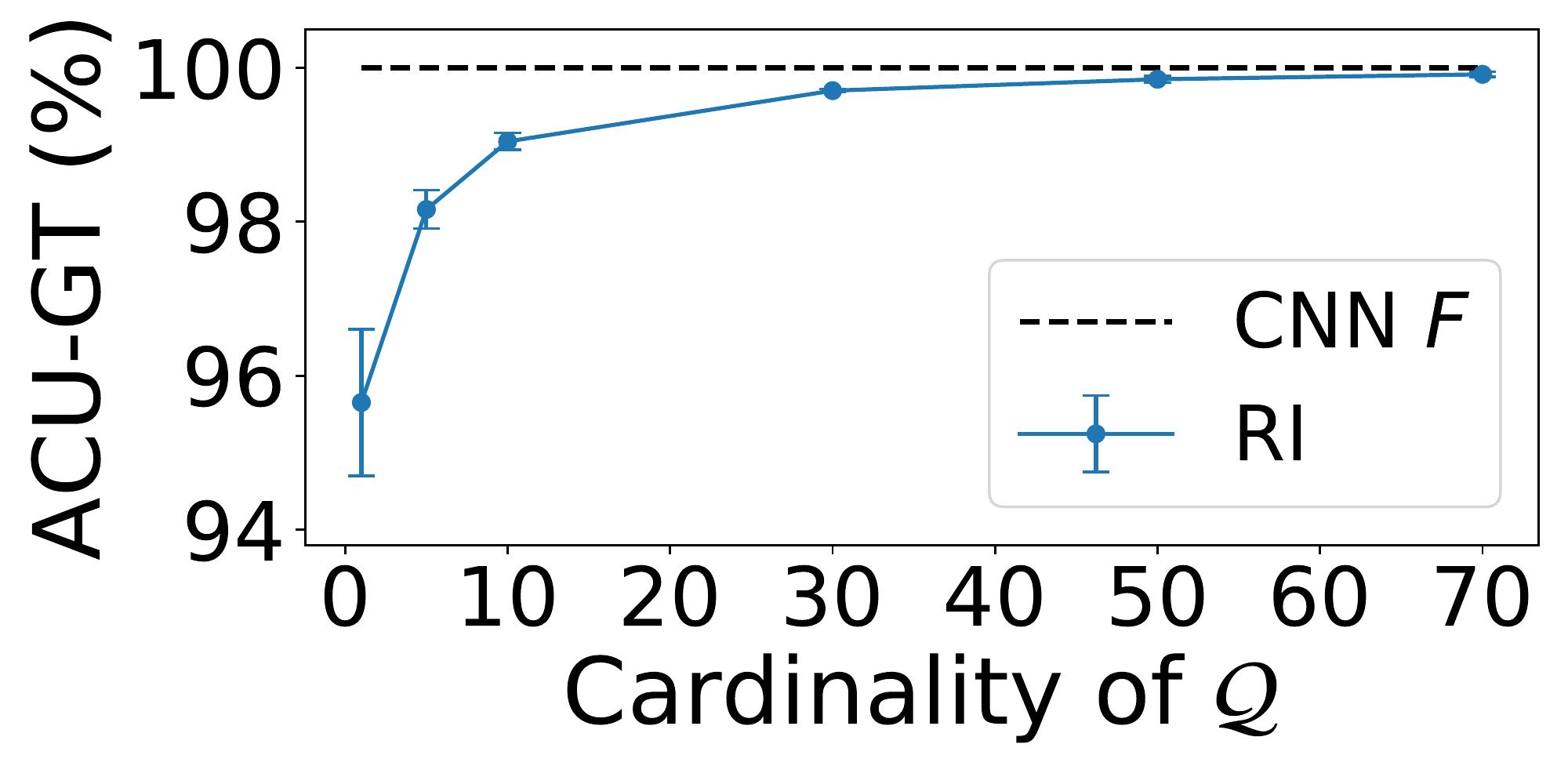}}
\subfigure[Unseen dataset of ASIRRA]{\includegraphics[width=0.22\textwidth]{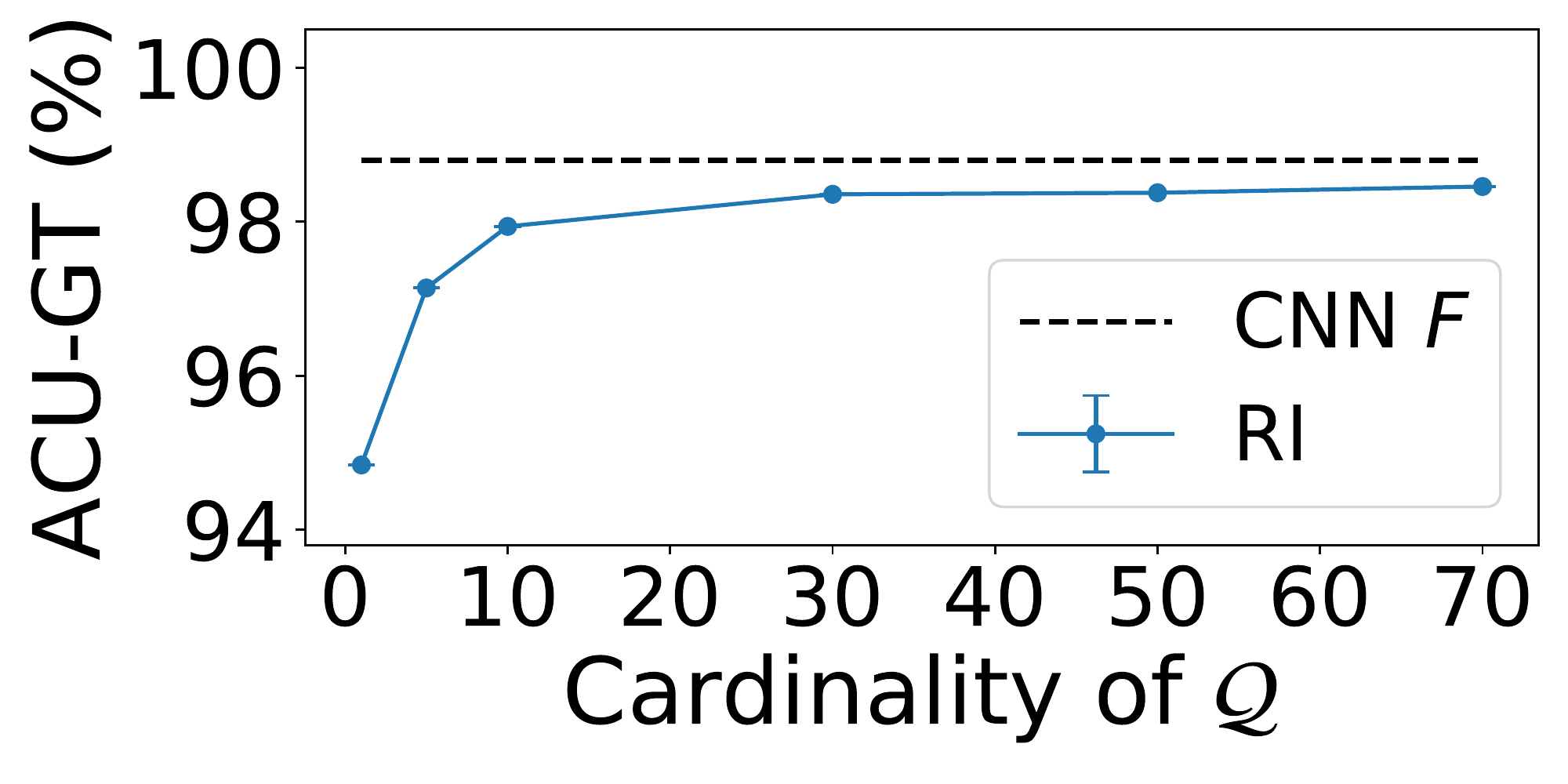}}
\newline
%===================================================
\subfigure[Reference dataset of GC]{\includegraphics[width=0.22\textwidth]{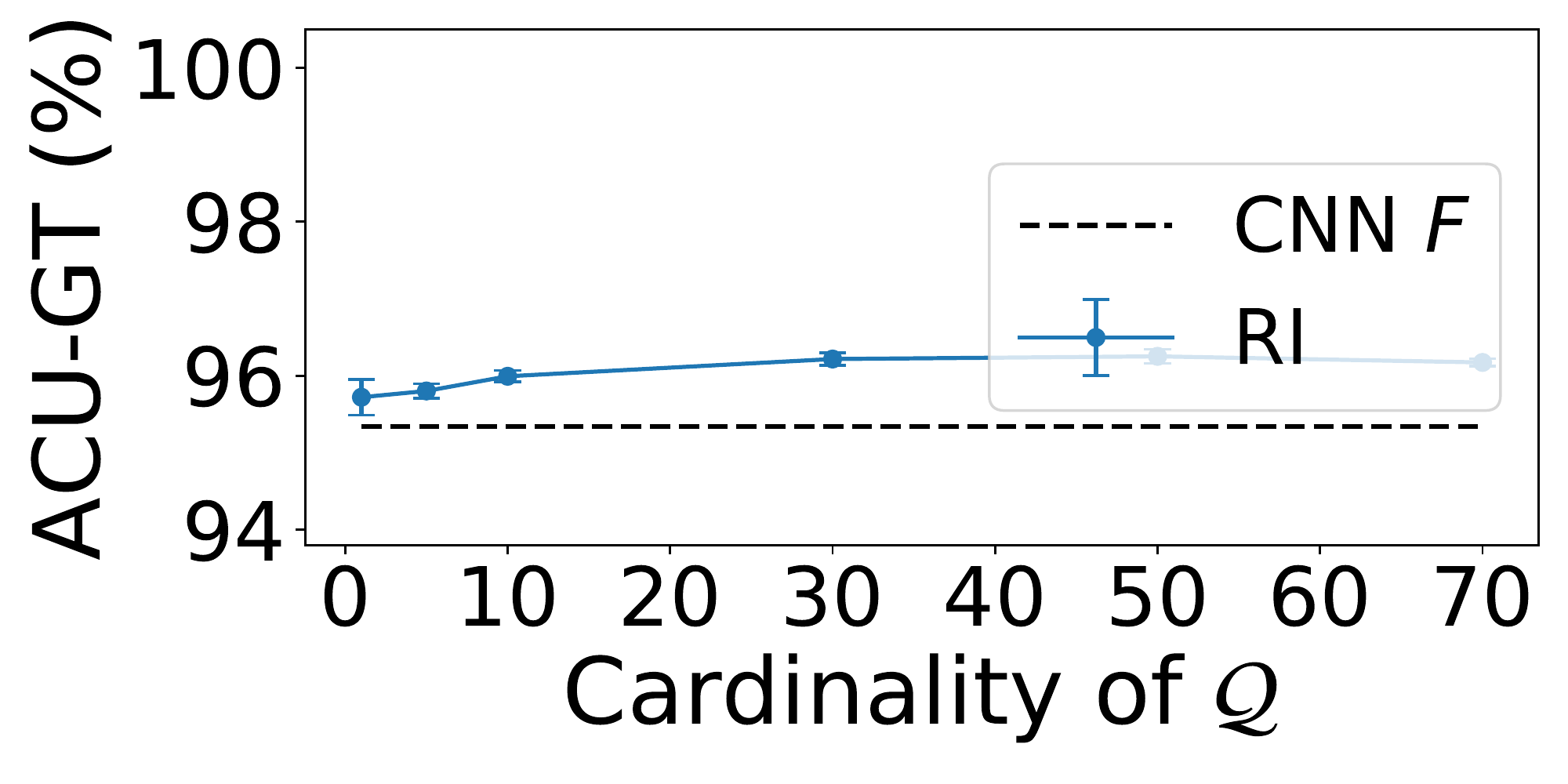}}
\subfigure[Unseen dataset of GC]{\includegraphics[width=0.22\textwidth]{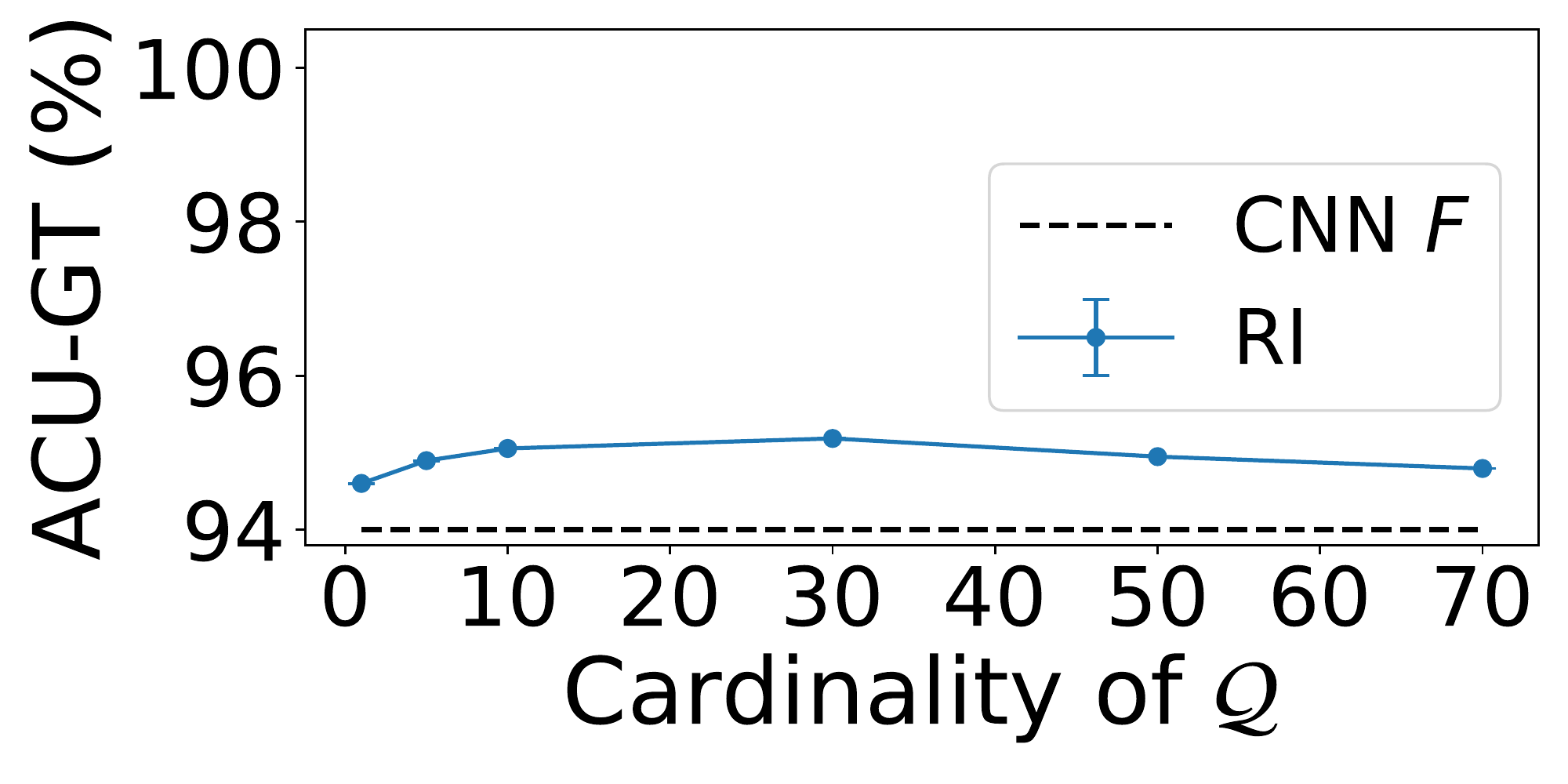}}
\newline
%===================================================
\subfigure[Reference dataset of RO]{\includegraphics[width=0.22\textwidth]{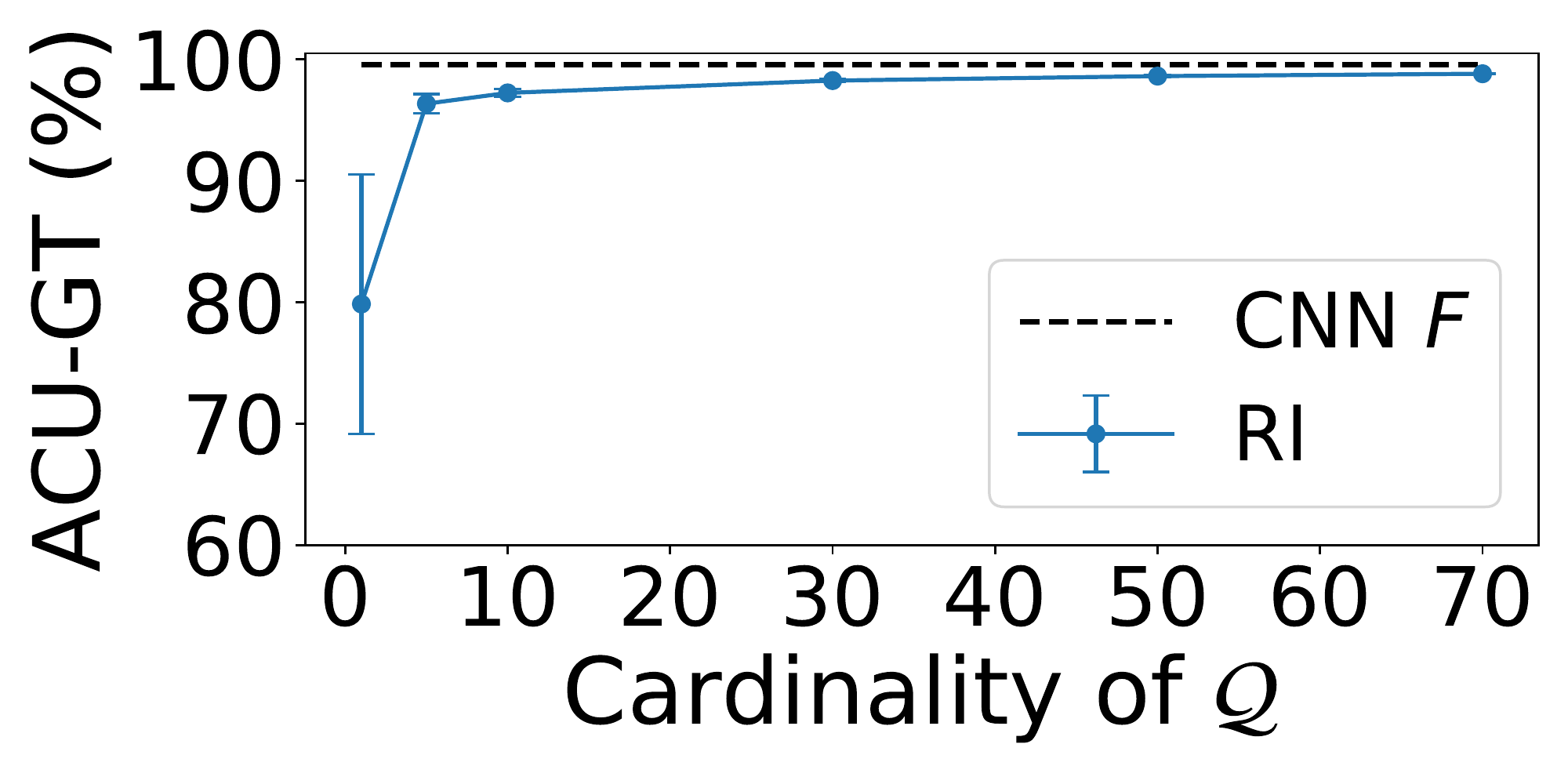}}
\subfigure[Unseen dataset of RO]{\includegraphics[width=0.22\textwidth]{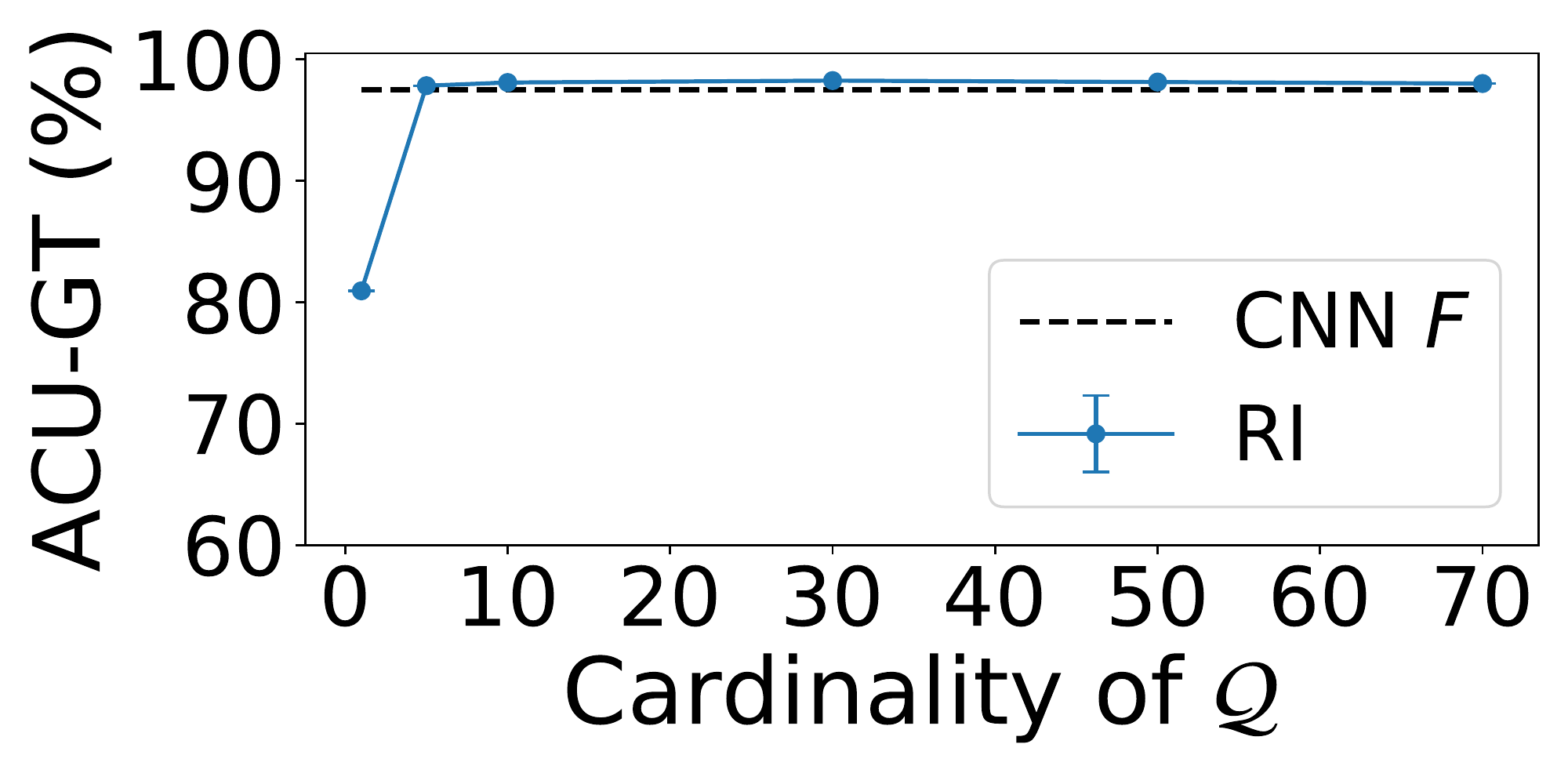}}
\newline
%===================================================
\subfigure[Reference dataset of FOOD]{\includegraphics[width=0.22\textwidth]{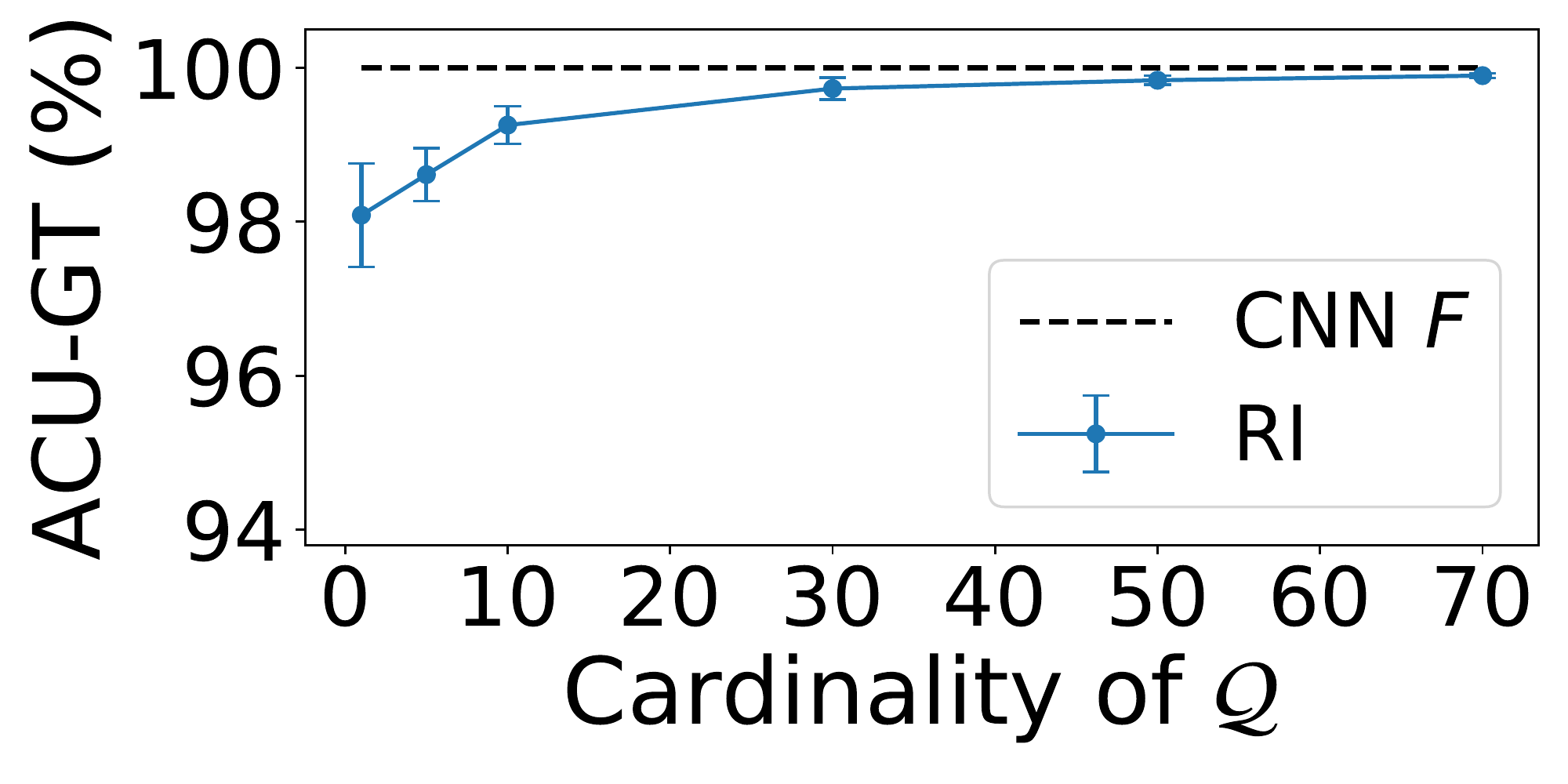}}
\subfigure[Unseen dataset of FOOD]{\includegraphics[width=0.22\textwidth]{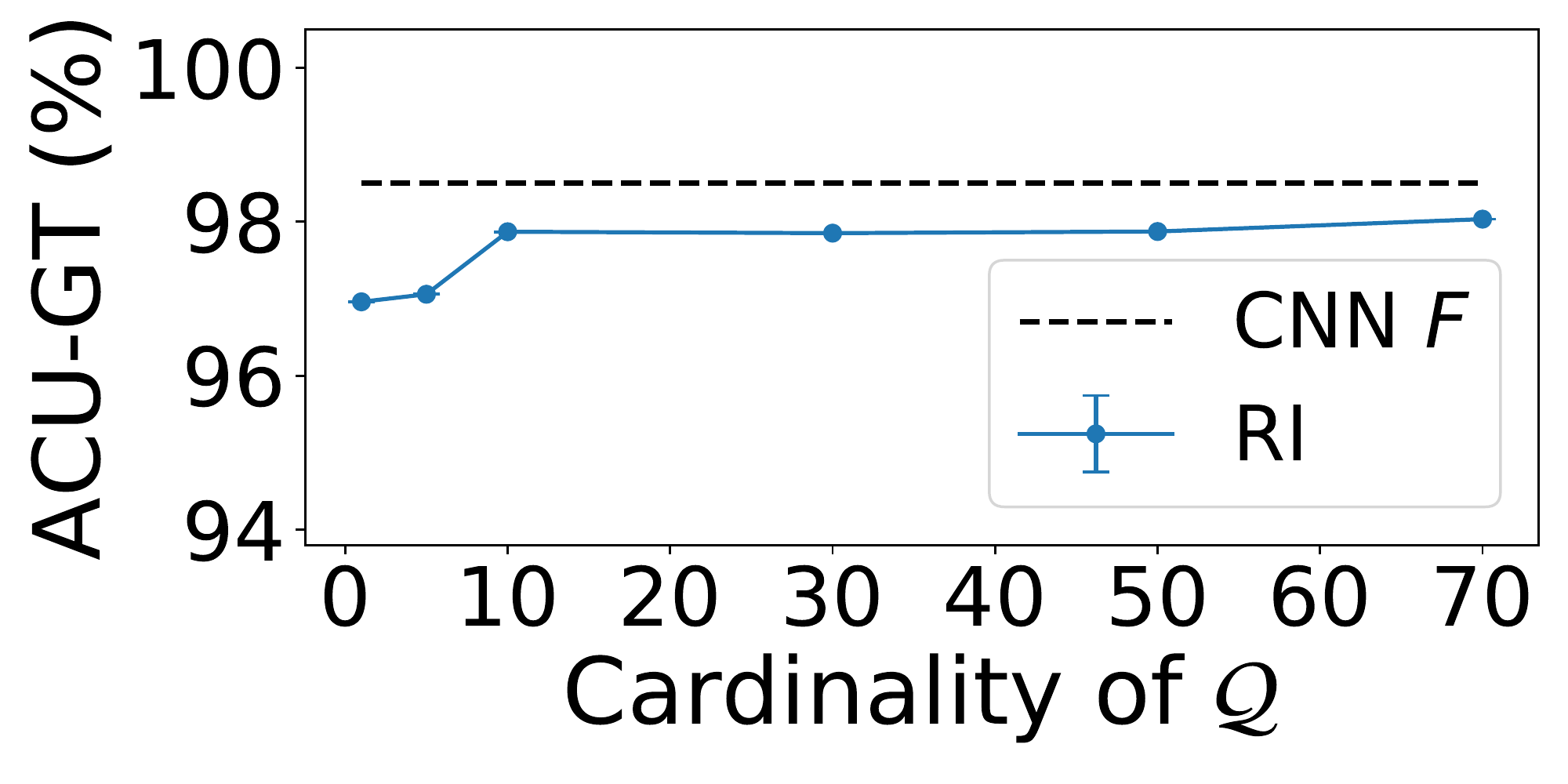}}
\newline
%===================================================
\caption{The ACU-GT performance of RI on the reference and unseen datasets of ASIRRA, GC, RO and FOOD.}
\label{fig:pa_accuracy}
\end{figure}

\begin{figure}[t]
\centering
%===================================================
\subfigure[Reference dataset of ASIRRA]{\includegraphics[width=0.22\textwidth]{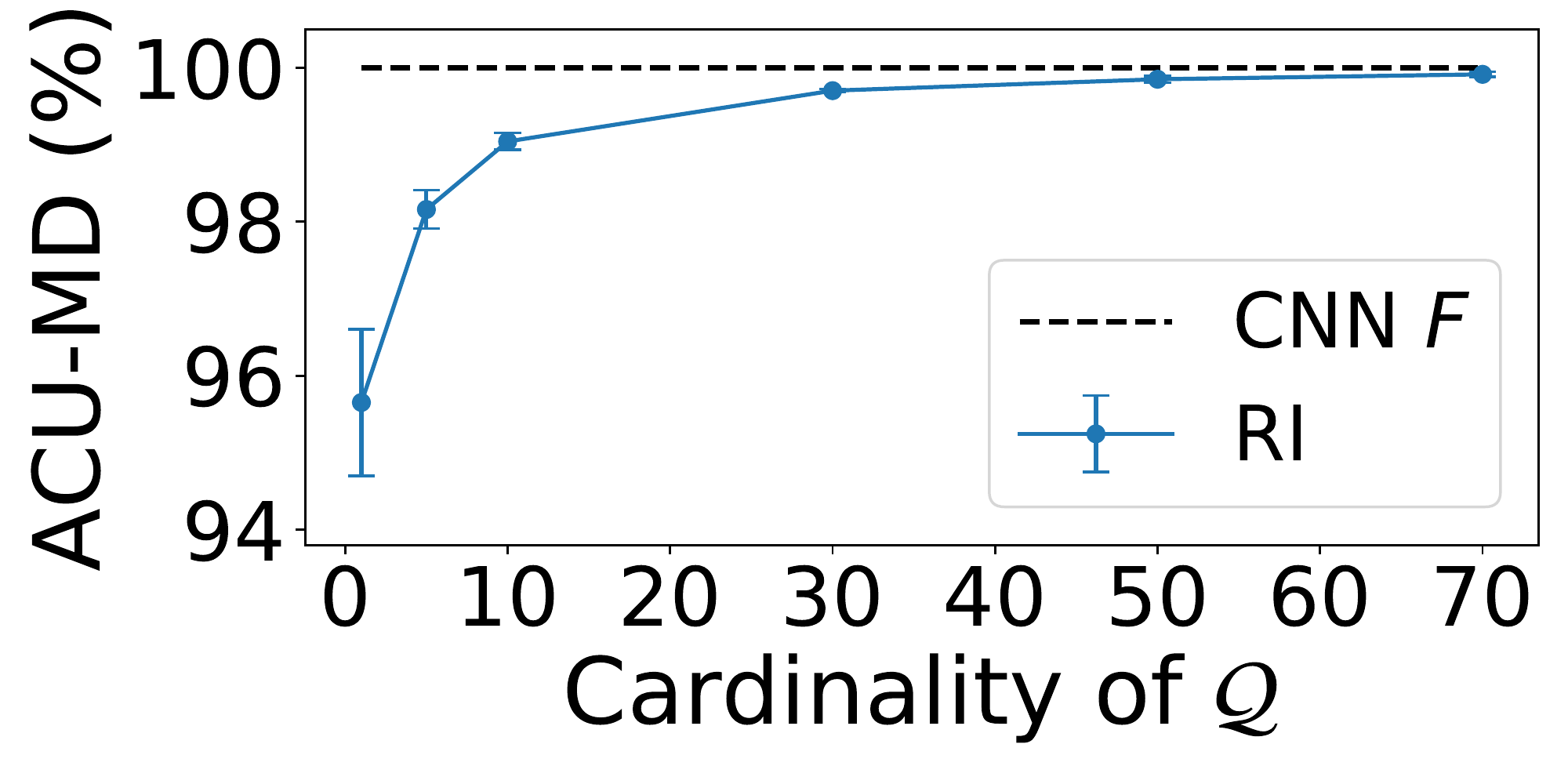}}
\subfigure[Unseen dataset of ASIRRA]{\includegraphics[width=0.22\textwidth]{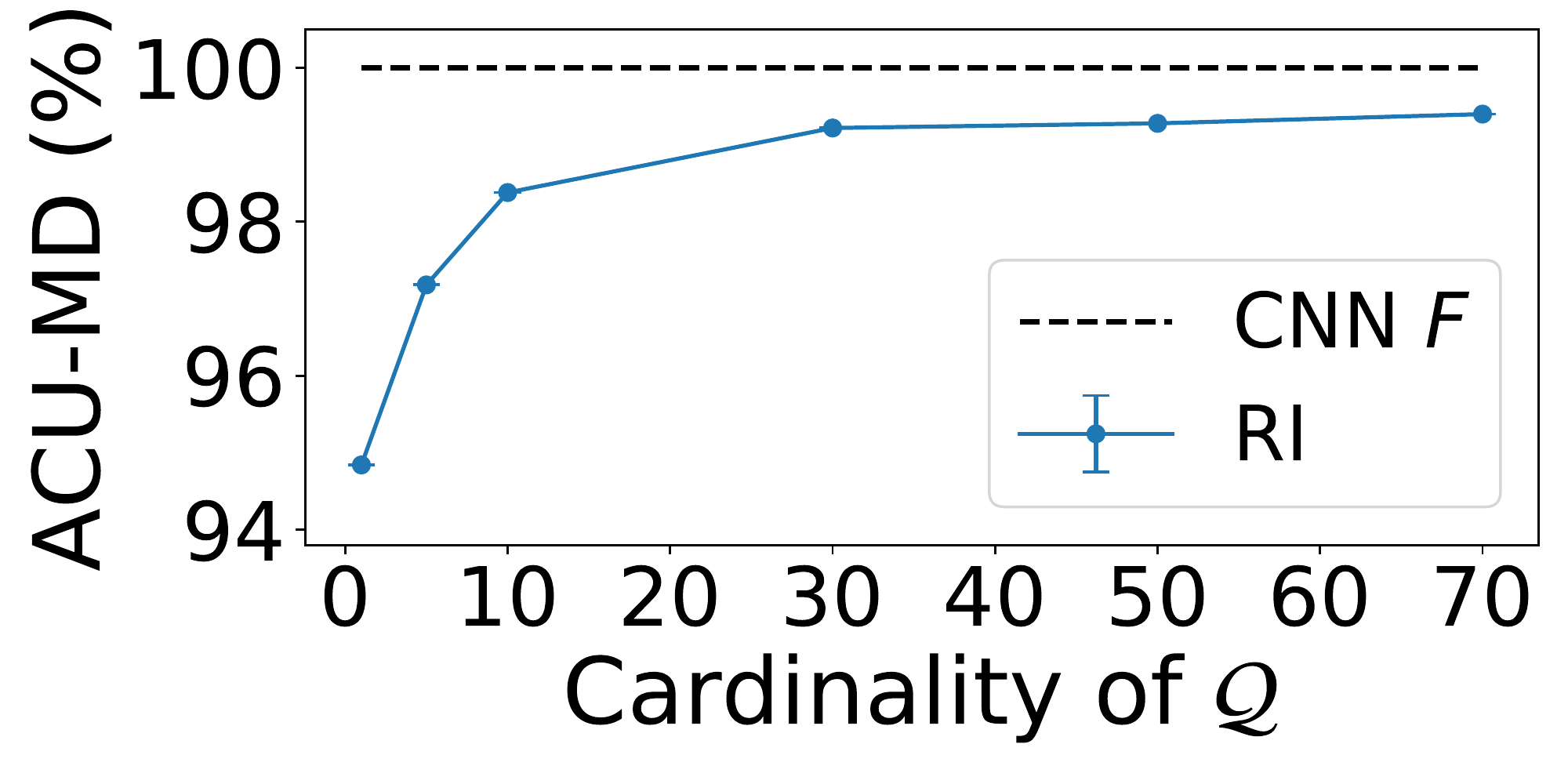}}
\newline
%===================================================
\subfigure[Reference dataset of GC]{\includegraphics[width=0.22\textwidth]{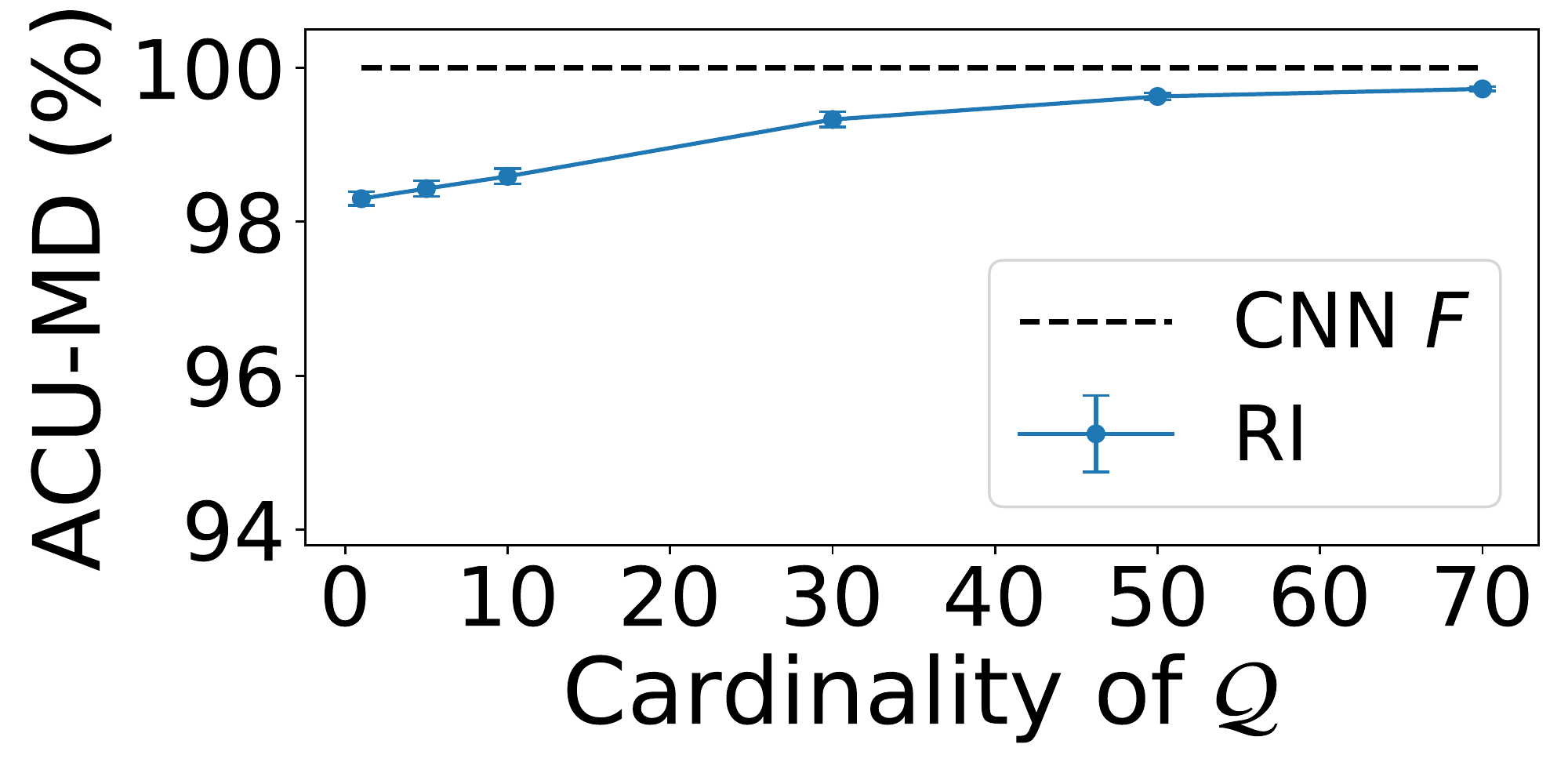}}
\subfigure[Unseen dataset of GC]{\includegraphics[width=0.22\textwidth]{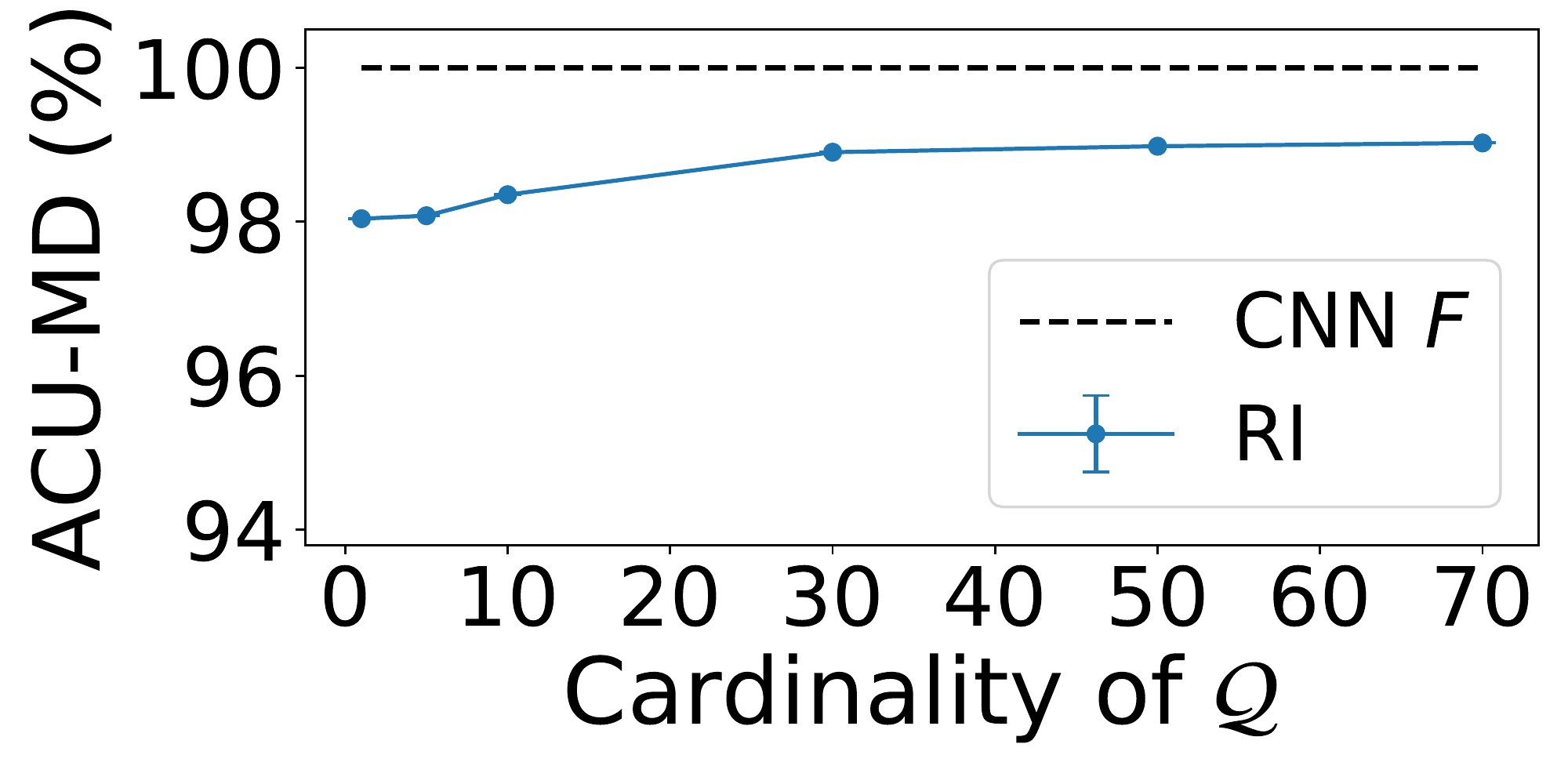}}
\newline
%===================================================
\subfigure[Reference dataset of RO]{\includegraphics[width=0.22\textwidth]{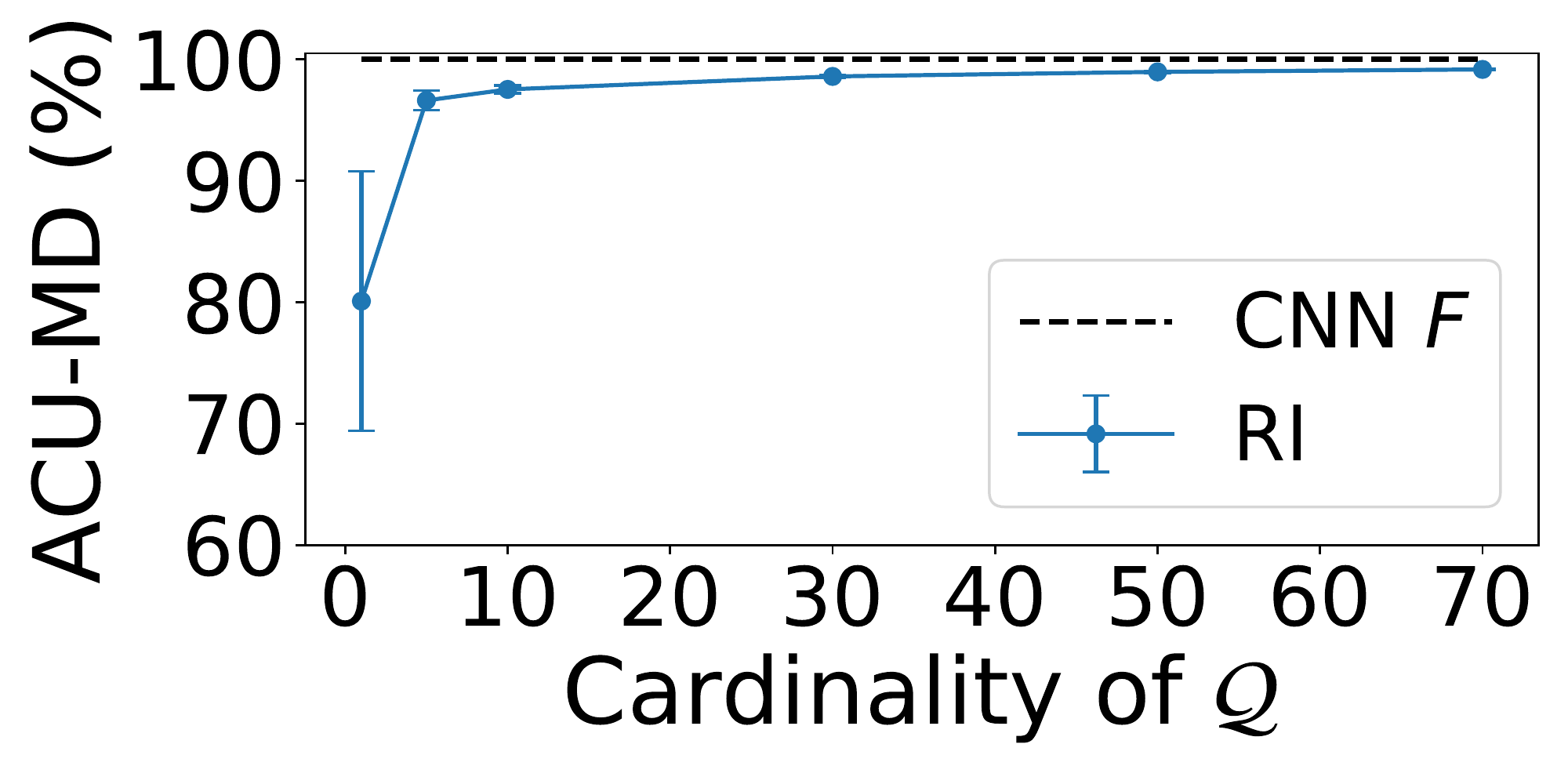}}
\subfigure[Unseen dataset of RO]{\includegraphics[width=0.22\textwidth]{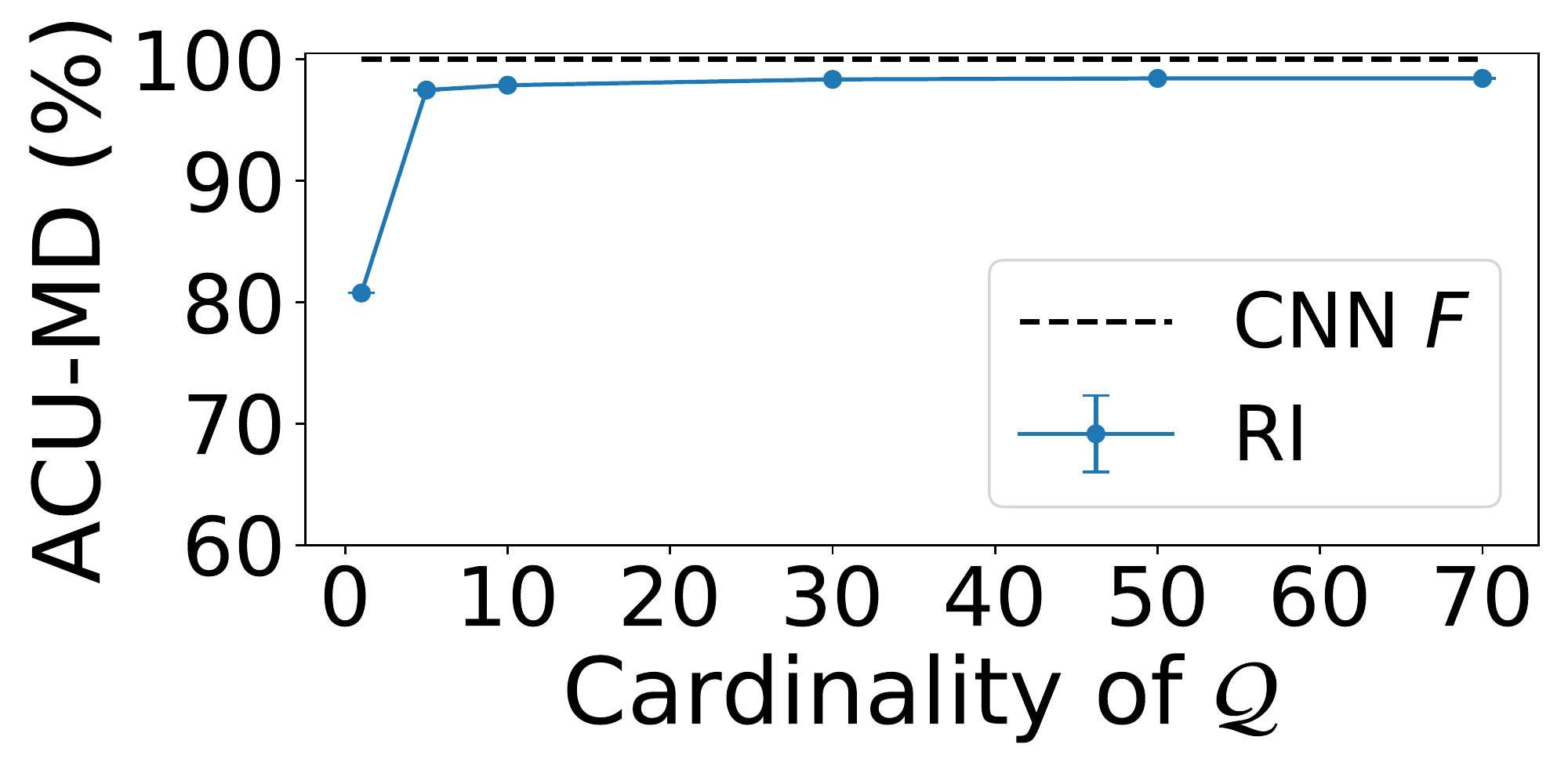}}
\newline
%===================================================
\subfigure[Reference dataset of FOOD]{\includegraphics[width=0.22\textwidth]{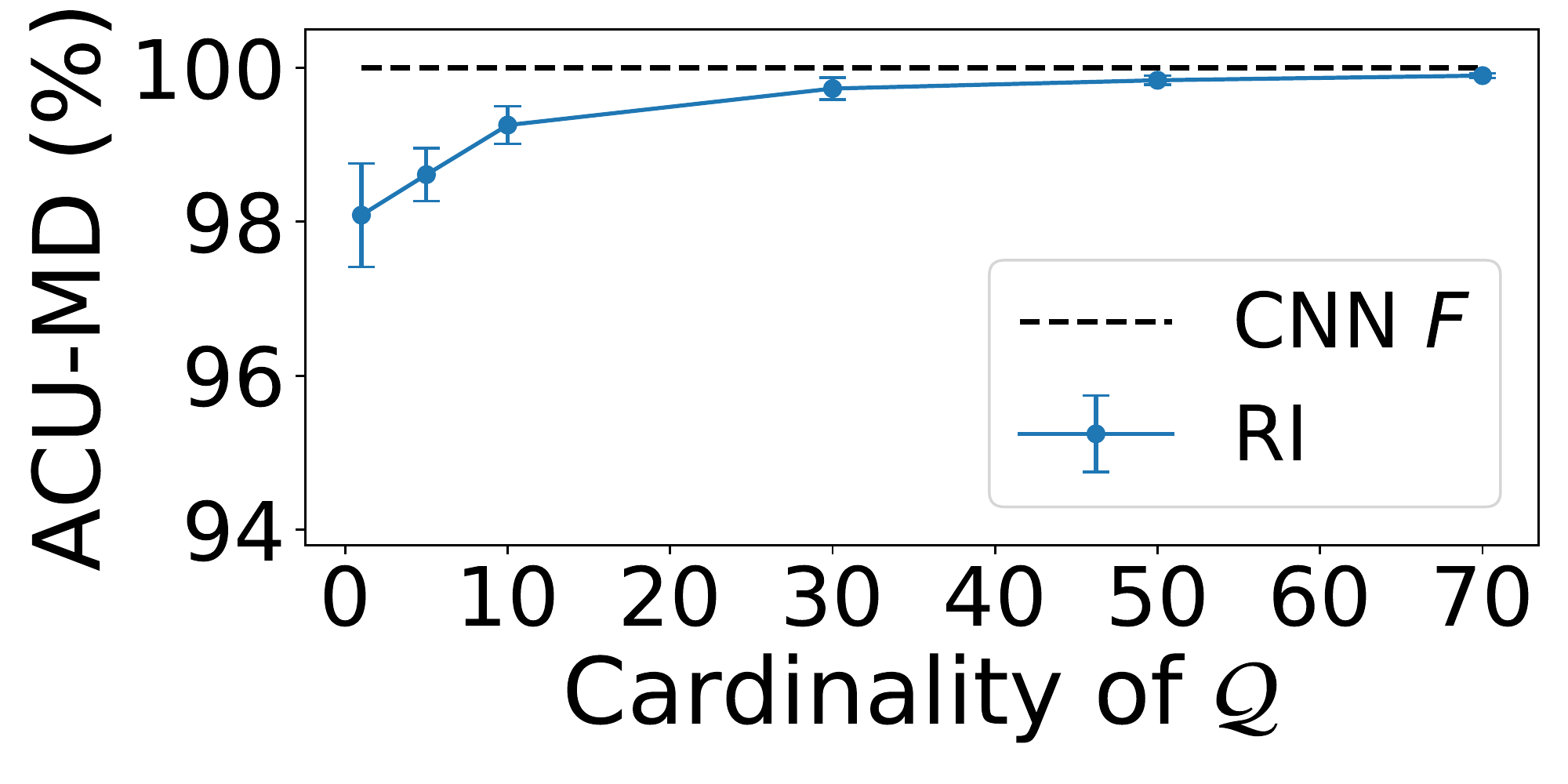}}
\subfigure[Unseen dataset of FOOD]{\includegraphics[width=0.22\textwidth]{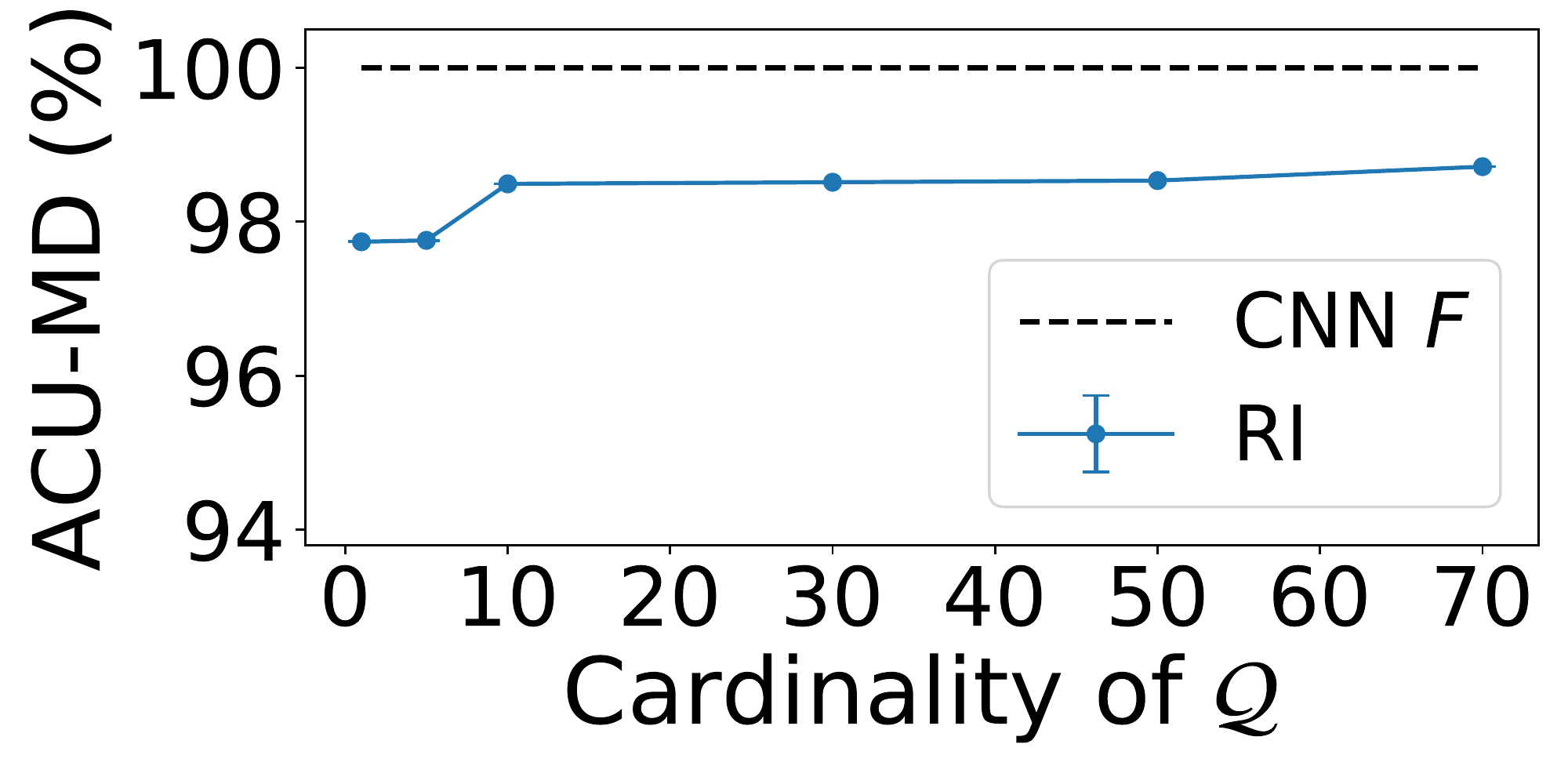}}
\newline
%===================================================
\caption{The ACU-MD performance on the reference and unseen datasets of ASIRRA, GC, RO and FOOD.}
\label{fig:pa_accuracy_wrt_model}
\end{figure}

\section{Parameter Analysis}

In this section, we analyze how the cardinality of $\mathcal{Q}$, denoted by $|\mathcal{Q}|$, affects the ACU-GT and ACU-MD performance of RI on every dataset.

We follow the same experiment setting as Appendix~\ref{sec:cov} to compute the mean and standard deviation of the ACU-GT and ACU-MD of RI for 6 different values of $|\mathcal{Q}|$, such as 1, 5, 10, 30, 50 and 70.

%For each value of $|Q|$, we run RI for 5 independent times, and compute the mean and standard deviation of the ACU-GT and ACU-MD performance, respectively.

Figure~\ref{fig:pa_accuracy} shows the ACU-GT of RI and the CNN model $F$ on the reference and unseen datasets of ASIRRA, GC, RO and FOOD.

Each solid point on the blue solid curve shows the mean ACU-GT of RI, and the corresponding error bar shows the standard deviation of the ACU-GT of RI.

The ACU-GT of $F$ is a single scalar for each value of $|\mathcal{Q}|$, thus we draw the ACU-GT of $F$ as a dashed curve without error bars.

Figure~\ref{fig:pa_accuracy_wrt_model} is drawn in a similar way as Figure~\ref{fig:pa_accuracy} to show the ACU-MD of RI and $F$ on each dataset. The ACU-MD of $F$ is always 100\% because the prediction results of $F$ is used as ground truth to compute ACU-MD.

We can see from Figure~\ref{fig:pa_accuracy} and Figure~\ref{fig:pa_accuracy_wrt_model} that 
the ACU-GT and ACU-MD of RI are small when $|\mathcal{Q}|$ is small. 
This is because every convex polytope produced by RI consists of some linear boundaries in $\mathcal{Q}$. 
If the number of linear boundaries in $\mathcal{Q}$ is too small, a convex polytope produced by RI will not have enough linear boundaries to approximate the complex decision logic of $F$. Therefore, the ACU-GT and ACU-MD of RI will be compromised.

We can also see that, when the cardinality of $\mathcal{Q}$ increases, the ACU-GT and ACU-MD of RI first increase and then become stable when $|\mathcal{Q}|$ is large.
The reason is that increasing the number of linear boundaries in $\mathcal{Q}$ largely improves the descriptive power of the convex polytopes produced by RI, thus the ACU-GT and ACU-MD of RI increases significantly in the beginning. 
However, when the number of linear boundaries in $\mathcal{Q}$ is large, existing linear boundaries in $\mathcal{Q}$ are good enough to produce high-quality convex polytopes, thus adding newly sampled linear boundaries into $\mathcal{Q}$ will not further increase the ACU-GT and ACU-MD of RI very much. In consequence, the ACU-GT and ACU-MD of RI become stable when $\mathcal{Q}$ is large.

Recall that a high ACU-GT indicates that the interpretations produced by RI capture some useful patterns of data to make accurate predictions; and a high ACU-MD means the interpretations produced by RI is close to the decision logic of $F$.
According to the results in Figure~\ref{fig:pa_accuracy} and Figure~\ref{fig:pa_accuracy_wrt_model}, since the ACU-GT and ACU-MD of RI are large and stable on all the datasets when $|\mathcal{Q}|$ is larger than 50, we simply set $|\mathcal{Q}|=50$ for RI to achieve outstanding interpretation performance in our experiments.

\section{A/B Test on Retina OCT Dataset}
\label{sec:abtest}

In this section, we conduct an A/B test on the Retina OCT (RO) dataset to demonstrate the effectiveness of RI in improving the diagnosis accuracy of human on retina disease.

The setting of the A/B test is as follows.

\textbf{Task}: we formulate a binary classification task using images from the two classes of NORMAL and DME in the RO dataset. NORMAL represents the images of normal retina and DME is a type of retina disease. The goal of the task is to predict whether an input image is NORMAL or DME.

\textbf{The CNN model and the interpretation method}: we train a VGG-19 model~\cite{Simonyan2015VeryDC} to achieve a testing accuracy of 97\% for the binary classification task between NORMAL and DME. We use RI to produce representative interpretations on the VGG-19 model.

\begin{figure}[t]
\centering
\includegraphics[width=55mm]{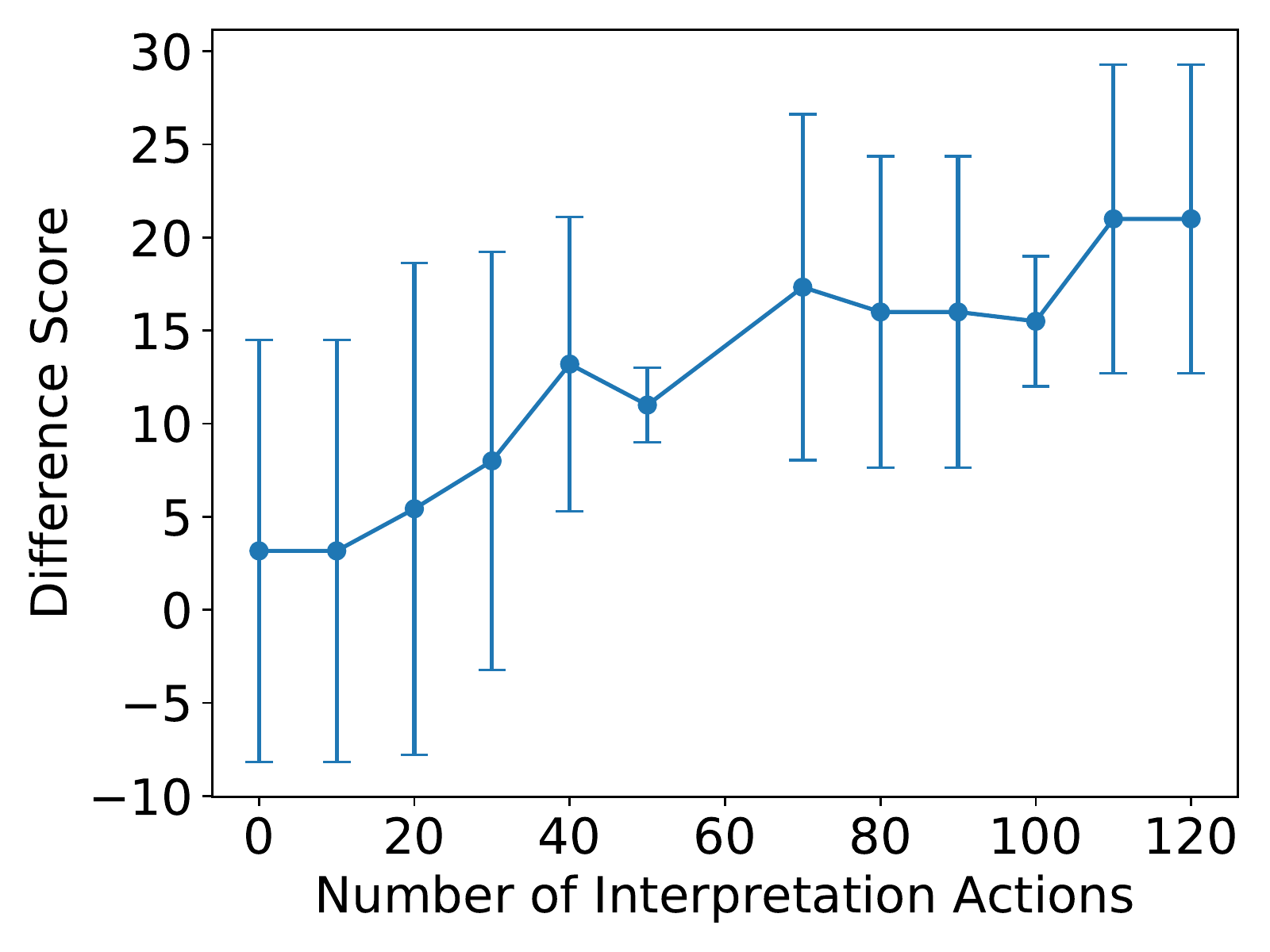}
\caption{The relationship between the difference score and the number of interpretation actions.
The $x$-axis represents the number of interpretation actions, which is between $0$ and $150$. 
The $y$-axis represents the difference score.
}
\label{fig:abtest}
\end{figure}

\begin{figure*}[t]
\centering
\includegraphics[width=172mm]{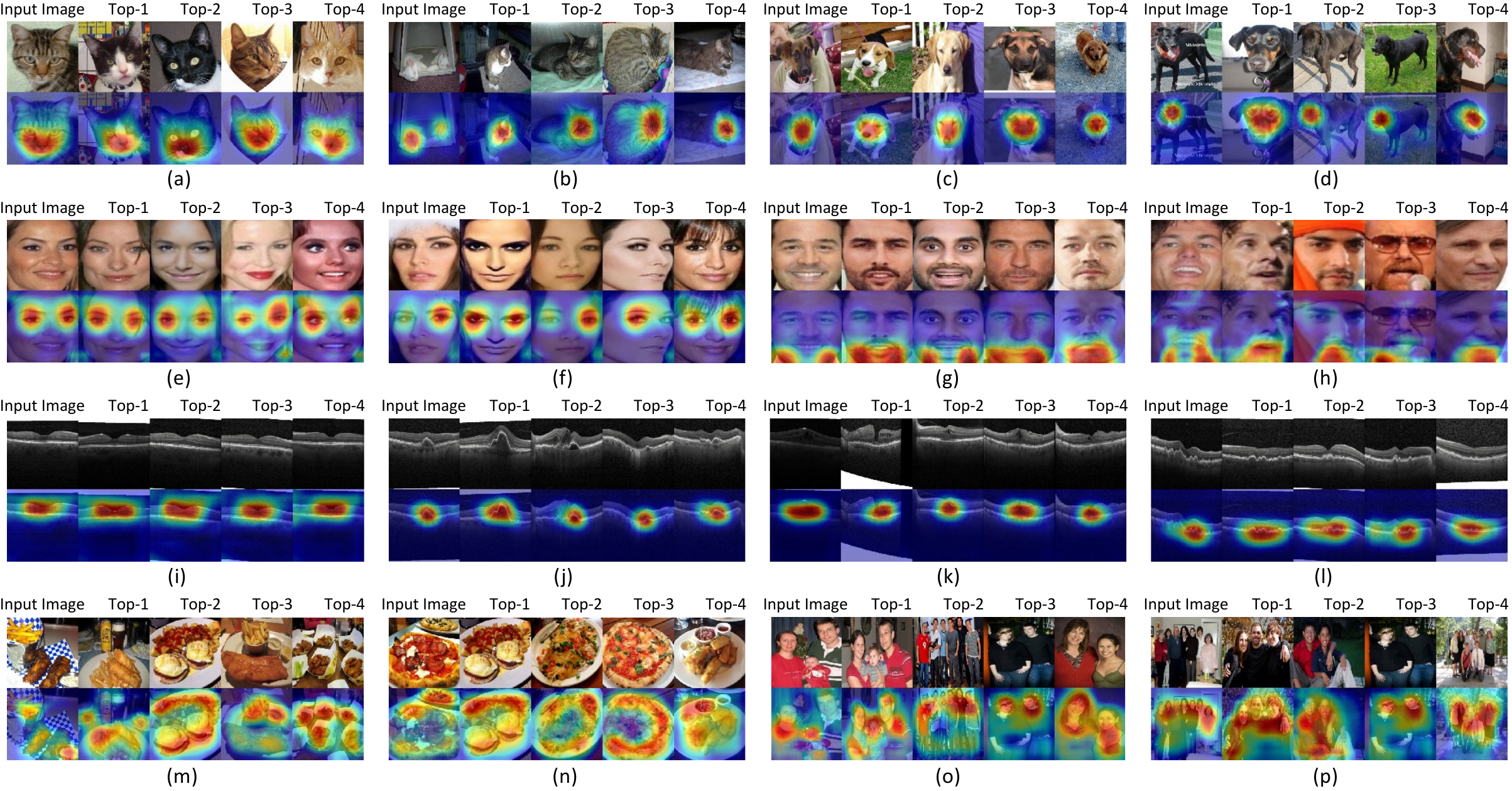}
\caption{Representative interpretations produced by RI on ASIRRA, GC, RO and FOOD.
For each figure, the first column is the input image and the rest of the columns are the similar images ranked by RI.
(a) and (b) show the interpretation results on two images of the class `cat' in ASIRRA. 
(c) and (d) show the interpretation results on two images of the class `dog' in ASIRRA.
(e) and (f) show the interpretation results on two images of the class `female' in GC.
(g) and (h) show the interpretation results on two images of the class `male' in GC.
(i), (j), (k) and (l) show the interpretation results on the images of the classes `NORMAL', `CNV', `DME' and `DRUSEN' in RO, respectively.
(m) and (n) show the interpretation results on two images of the class `food' in FOOD. 
(o) and (p) show the interpretation results on two images of the class `non-food' in FOOD.
}
\label{fig:extra_examples}
\end{figure*}

\textbf{Subjects}: the subjects of the A/B test is a group of 20 people without any knowledge background on retina disease.
To prepare these people for the binary classification task, we give the same written tutorial to each subject to teach them the basic skills to distinguish between NORMAL and DME.
Every subject has 10 minutes to read the tutorial.

\textbf{Test A}: the test A consists of 50 multiple-choices questions. Each question requires the subject to answer whether a retina image is NORMAL or DME. The choices of answers are: `Definite DME', `Maybe DME', `Not sure', `Maybe NORMAL', and `Definite NORMAL'. 
A `Not sure' answer receives a score of 0.
For answers with `Maybe', a correct one receives a score of +1, but a wrong one receives a score of -1.
For answers with `Definite', a correct one receives a score of +2, but a wrong one receives a score of -2.
Every subject taking test A will see the prediction results of the CNN model on the input retina images.
However,
to simulate the practical scenario that many people may not trust a machine learning model in making diagnosis, we lie to every subject that the CNN model only has a testing accuracy of 80\%.
Every subject is taking the same set of 50 questions, but the order of the questions is randomly generated for each subject.
The sum of the scores of the 50 questions are collected as the final score for each subject in test A.

\textbf{Test B}: the test B follows exactly the setting of test A.
The only difference is that a subject is able to see the interpretation result generated by RI for each input retina image.
We allow a subject to freely choose whether or not to see the interpretation result for the input image of each question.
We log three types of \textbf{interpretation actions} when a subject sees the interpretation result; these actions include: (i) `show similar images' shows the similar retina images to the input image ranked by RI; (ii) `show heat map' shows the heat map generated by RI on the input retina image and the similar retina images; and (iii) `Zoom in' enlarges the similar images to show more details.
The sum of the scores of the 50 questions are collected as the final score for each subject in test B.

Every subject is required to take Test A first and then take test B.
Since the orders of the 50 questions are randomly generated for every test, it is very difficult for a subject to memorize his/her answers in test A when taking test B.

Every subject produces a final score for test A, denoted by `Score A', and a final score for test B, denoted by `Score B'.
We collect the \textbf{difference score} between Score B and Score A, that is, Score B minus Score A, for each subject.
This produces 20 difference scores.

A larger difference score means Score B is higher than Score A, which indicates using interpretations produced by RI can improve the diagnosis accuracy of human on retina disease.

Recall that we also log the interpretation actions of each subject in test B. We collect the total number of interpretation actions of each subject to measure how often a subject uses the interpretation results produced by RI.

We draw the results produced by the 20 subjects as the blue points and the corresponding error bars in Figure~\ref{fig:abtest}.
Denote by $(x, y)$ the coordinates of a blue point, $y$ is the mean of the difference scores of all the subjects whose numbers of actions fall into the interval of $(x, x+30]$.
The corresponding error bar shows the standard deviation of the difference scores of the subjects.
There is no blue point for $x=60$ since there is no participant whose number of interpretation actions is between 60 and 90.

Figure 5 shows that a more frequent use of the interpretation results produced by RI contributes to a higher difference score.
This demonstrates the high effectiveness of RI in helping people making more accurate diagnosis on retina images.

\section{More Interpretation Examples of RI}
\label{sec:extra}
We present more examples in Figure~\ref{fig:extra_examples} to show the good interpretation performance of RI on each of ASIRRA, GC, RO and FOOD.

We can see from the results that the representative interpretations produced by RI always highlight meaningful common parts of the input image and the similar images.
For example, the faces of cats and dogs in Figures~\ref{fig:extra_examples}(a)-\ref{fig:extra_examples}(d), and the beard of male in Figures~\ref{fig:extra_examples}(g) and \ref{fig:extra_examples}(h).

Obviously, showing similar images with common highlighted parts as the input image makes our interpretations more convincing than showing only the interpretation on the input image.

The results in Figure~\ref{fig:extra_examples} further demonstrate the outstanding performance of RI in producing representative interpretations to reveal the common decision logic of a CNN on an input image as well as the images that are similar to the input image.

\end{document}